\documentclass{article}

\usepackage{arxiv}

\usepackage[utf8]{inputenc} 
\usepackage[T1]{fontenc}    
\usepackage{hyperref}       
\usepackage{url}            
\usepackage{booktabs}       
\usepackage{amsfonts}       
\usepackage{nicefrac}       
\usepackage{microtype}      
\usepackage{lipsum}		
\usepackage{graphicx}
\usepackage{natbib}
\usepackage{doi}

\usepackage{amsmath}
\usepackage{amssymb}
\usepackage{algorithm}
\usepackage{algpseudocode}
\usepackage{tikz}
\usepackage{graphicx,subfigure}
\usepackage{makecell}
\usetikzlibrary{shapes.arrows}
\usetikzlibrary{calc}
\usepackage{array,multirow}
\DeclareMathOperator*{\minimize}{minimize}
\DeclareMathOperator*{\argmin}{argmin}
\usepackage{multicol}
\usetikzlibrary{arrows.meta}
\usepackage{xcolor,colortbl}
\definecolor{green}{rgb}{0.1,0.1,0.1}
\newcommand{\done}{\cellcolor{gray!25}}  
\newcommand{\happy}{\cellcolor{yellow!25}}  
\usepackage{makecell}

\title{High-fidelity Interpretable Inverse Rig: An Accurate and Sparse Solution Optimizing the Quartic Blendshape Model}

\date{February 2023}	

\author{ \href{https://orcid.org/0000-0002-5656-9189}{\includegraphics[scale=0.06]{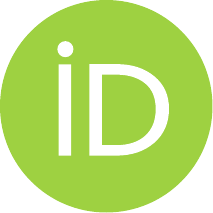}\hspace{1mm}Stevo Racković} \\
	Department of Mathematics\\
	Instituto Superior Técnico\\
	Lisbon, Portugal \\
	\texttt{stevo.rackovic@tecnico.ulisboa.pt} \\
	\And
	\href{https://orcid.org/0000-0003-3071-6627}{\includegraphics[scale=0.06]{orcid.pdf}\hspace{1mm}Cláudia Soares} \\
	Department of Computer Sceince\\
	NOVA School of Science and Technology\\
	Caparica, Portugal \\
	\And
	\href{https://orcid.org/0000-0003-3497-5589}{\includegraphics[scale=0.06]{orcid.pdf}\hspace{1mm}Dušan Jakovetić} \\
	Department of Mathematics\\
	University of Novi Sad\\
	Novi Sad, Serbia \\
 	\And
	Zoranka Desnica \\
	3Lateral Animation Studio\\
	Epic Games Company \\
}

\renewcommand{\shorttitle}{ }

\hypersetup{
pdftitle={High-fidelity Interpretable Inverse Rig: An Accurate and Sparse Solution Optimizing the Quartic Blendshape Model},
pdfauthor={Stevo Racković, Cláudia Soares, Dušan Jakovetić, Zoranka Desnica},
pdfkeywords={Inverse Rig, Quadratic blendshape model, Majorization Minimization},
}

\begin{document}
\maketitle

\begin{abstract}
    We propose a method to fit arbitrarily accurate blendshape rig models by solving the inverse rig problem in realistic human face animation. The method considers blendshape models with different levels of added corrections and solves the regularized least-squares problem using coordinate descent, i.e., iteratively estimating blendshape weights. Besides making the optimization easier to solve, this approach ensures that mutually exclusive controllers will not be activated simultaneously and improves the goodness of fit after each iteration. We show experimentally that the proposed method yields solutions with mesh error comparable to or lower than the state-of-the-art approaches while significantly reducing the cardinality of the weight vector (over $20\%$), hence giving a high-fidelity reconstruction of the reference expression that is easier to manipulate in the post-production manually. Python scripts for the algorithm will be publicly available upon acceptance of the paper. 
\end{abstract}



\section{Introduction}\label{sec:intro}

\begin{figure}
          \begin{tikzpicture}
        \node[above right, inner sep=0] (image) at (0,2.4){\includegraphics[width=\textwidth]{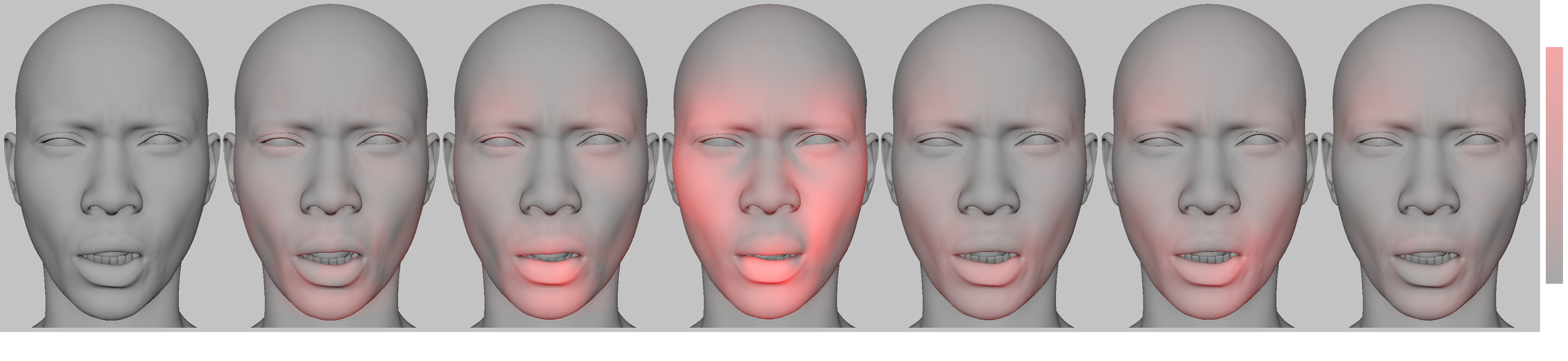}};
        \node[above right, inner sep=0] (image) at (-0.2,0){\includegraphics[width=\textwidth]{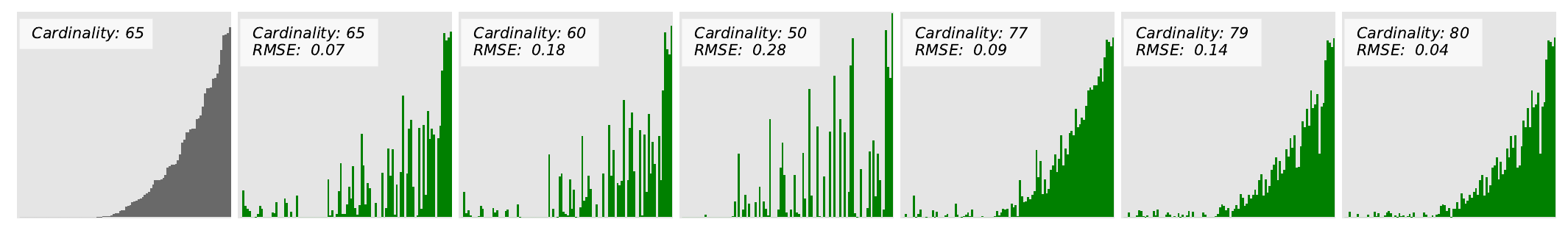}};
        \begin{scope}[
        x={($0.1*(image.south east)$)},
        y={($0.1*(image.north west)$)}]
            \node[darkgray] at  (0.70,0.2){\small Reference };
            \node[darkgray] at  (2.10,0.2){\small Quartic (ours) };
            \node[darkgray] at  (3.50,0.2){\small Linear (ours) };
            \node[darkgray] at  (4.95,0.2){\small Seol };
            \node[darkgray] at  (6.30,0.2){\small Joshi };
            \node[darkgray] at  (7.80,0.2){\small Çetinaslan };
            \node[darkgray] at  (9.20,0.2){\small LMMM };
            \node[darkgray] at  (10.35,23.5){\scriptsize .29 };
            \node[darkgray] at  (10.15,12){\scriptsize .00 };
            \node[darkgray] at  (10.15,11){\scriptsize cm };
        \end{scope}
        \end{tikzpicture}
    \caption{\textit{Ada}, an example frame with predictions using different methods. The top row shows a mesh reconstruction, and regions of higher mesh error are highlighted in red, according to the color bar on the right. The bottom row shows corresponding blendshape weights activation, with summarized root mean squared error and cardinality of each approach. The method of \textit{Seol} \cite{seol2011artist} gives the sparsest vector of weights, yet  the error is considerably higher than for the other approaches, and the resulting expression is wrong. Our approach with linear rig approximation (\textit{Linear}) corrects some of these artifacts, yet the lips still do not fit well enough, similar to \textit{Joshi} \cite{joshi2006learning} and \textit{Çetinaslan} \cite{cetinaslan2020sketching}. The solution of \textit{LMMM} \cite{rackovic2022majorization} leads to the lowest mesh error, but it activates the highest number of blendshapes. Only our approach with quartic corrective terms (\textit{Quartic}) gives a good trade-off between the accurate mesh fit and a low number of activated weights.}
  \label{fig:wide1}
\end{figure}
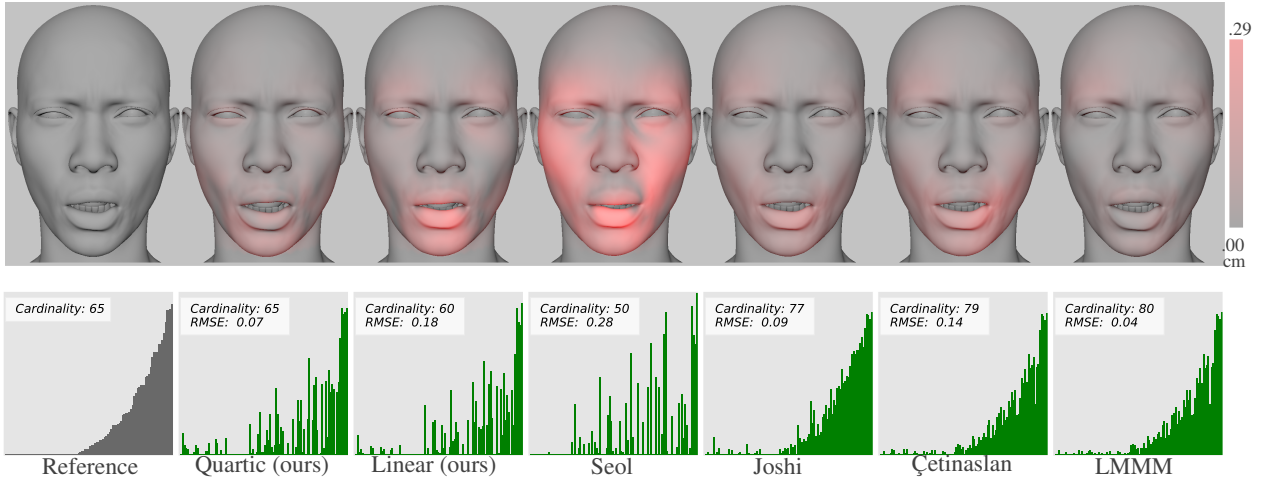

The human face has always occupied a central place in the animation industry due to its role in nonverbal communication and our high sensitivity to subtle expression changes. With the advances in movie and video game production, the models for representing a face are increasingly more complex and demand algorithms with ever-increasing accuracy and attention to detail in order to provide a high-fidelity appearance of the avatars. One of the most popular approaches for animating the face is the blendshape model \cite{lewis2014practice}. The main building blocks in the blendshape model are a neutral mesh, represented in a vector form as $\textbf{b}_0\in\mathbb{R}^{3n}$, where $n$ is the total number of vertices in the face, and $m$ blendshape vectors $\textbf{b}_1,...,\textbf{b}_m\in\mathbb{R}^{3n}$ that represent meshes corresponding to the atomic deformations of the face, and are later combined to build more complex expressions. In the case of the \textit{delta} blendshape model, these vectors are not actual meshes, but their offsets from the neutral, and they are added on top of the neutral face with corresponding activation weights $w_1,...,w_m$. A linear delta blendshape model is given by
\begin{equation}\label{eq:linear_blendshape_model}
    f_L(\textbf{w}) = \textbf{b}_0 +  \sum_{i=1}^m w_i\textbf{b}_i = \textbf{b}_0 + \textbf{Bw},
\end{equation}
where the subscript $L$ stands for \textit{linear}, in order to distinguish it from a more complex rig function that we will introduce later in this section. The matrix $\textbf{B}\in\mathbb{R}^{3n\times m}$ is a blendshape matrix whose columns are the above introduced blendshape vectors, and $\textbf{w}\in\mathbb{R}^m$ is a vector of activation weights $\textbf{w} = [w_1,...,w_m]^T$.

The problem that is of interest to us is the inversion of the rig to obtain activations $\textbf{w}$ from an acquired mesh, i.e., the inverse rig problem. It takes a given face mesh $\widehat{\textbf{b}}\in\mathbb{R}^{3n}$ as input and estimates a weight vector $\textbf{w}$ that would produce a good approximation to the mesh, i.e., such that $f_L(\textbf{w})\approx \widehat{\textbf{b}}$. This is often formulated as a least-squares fitting problem. In \cite{joshi2006learning}, the authors formulate the corresponding optimization problem as
\begin{equation}\label{eq:psd_problem}
    \minimize_{\textbf{w}} \frac{1}{2}\| f_L(\textbf{w}) - \widehat{\textbf{b}} \|^2,
\end{equation}
where $\|\cdot\|$ denotes the $l_2$ norm. A solution of (\ref{eq:psd_problem}) can be found in a closed form as 
\begin{equation}\label{eq:psd_solution}
    \textbf{w} = (\textbf{B}^T\textbf{B})^{-1}\textbf{B}^T\widehat{\textbf{b}} = \textbf{B}^{\dagger}\widehat{\textbf{b}},
\end{equation}
where $\textbf{B}^{\dagger}$ represents the pseudoinverse of the matrix $\textbf{B}$, and throughout the paper, we refer to this approach as \textit{Joshi}. This solution might be undesirable as it activates all the controllers, or a matrix $\textbf{B}^T\textbf{B}$ might be singular if the number of vertices is too few relative to the number of model weights \cite{lewis2010direct}; hence, more recent papers include additional regularization terms to problem (\ref{eq:psd_problem}). Another issue is that \textit{Joshi} does not incorporate constraints to the controllers --- for practical applications, weights are restricted to be non-negative since negative weights defy the intended semantics of the blendshapes and break the intuition for manual corrections \cite{lewis2014practice}. Similarly, weights should not exceed a value of 1, although some authors allow for this, justifying it as a way to produce exaggerated cartoonish expressions \cite{choe2001analysis}. To satisfy the constraints, the solution vector of \textit{Joshi} can be clipped to a feasible set, although this was not done in the original paper. Despite the above issues, the solution of \cite{joshi2006learning} is often encountered in the literature because of its simplicity. There is a range of variants of this approach --- versions in \cite{choe2001analysis, Liu2010LocalizedOF} include constraints $\textbf{0}\leq\textbf{w}\leq \textbf{1}$; in \cite{lewis2010direct, ccetinaslan2016position, cetinaslan2020sketching} the weights are regularized by a squared $l2$ norm to impose the stability of the solution, while the $l1$ norm was used in  \cite{ribera2017facial} to stress on the sparsity. It is common for all these models to estimate all the weights jointly. While this can sometimes lead to a more precise mesh fit, such an approach does not allow so much flexibility in manually adjusting the solution weights afterward. 

A solution proposed in \cite{seol2011artist} follows a different logic, relying on a two-step approach, motivated by the way artists animate manually. The weights are visited and estimated sequentially, updating the residuals after each controller. Initially, residual vector $\textbf{r}\in\mathbb{R}^{3n}$ is set to be equal to a given mesh $\textbf{r}^{(1)}:=\widehat{\textbf{b}}$. Then each of the $m$ controllers is visited once, solving the following two-step problem:
\begin{equation}\label{eq:seol_iteration}
   \begin{split}
       \text{For } i = 1,...,m: & \\
       \text{step 1: } & w_i \leftarrow \argmin_{0\leq w_i} \|\textbf{r}^{(i)}-\textbf{b}_iw_i\|^2, \\
       \text{step 2: } & \textbf{r}^{(i+1)} \leftarrow \textbf{r}^{(i)}-\textbf{b}_iw_i.
   \end{split}
\end{equation}
After each iteration $i$, \textit{step 1} finds an optimal weight for the blendshape $\textbf{b}_i$, and \textit{step 2} removes the corresponding estimated effect, producing a new residual vector $\textbf{r}^{(i+1)}$. The output is a vector of weights $\textbf{w}$, estimated in a greedy search manner to fit the original target mesh $\widehat{\textbf{b}}$.
Throughout the paper, we refer to this approach as \textit{Seol}.

An important advantage of \textit{Seol} is that it empirically avoids simultaneous activation of mutually exclusive blendshapes, which is one of the leading causes of instability in the solution. As confirmed experimentally in \cite{seol2011artist}, this also leads to a sparser solution, and hence it is easier to manipulate the animation later if needed. In this approach, the order in which controllers are visited plays an important role. The authors suggest ordering them by the magnitude of change each blendshape produces when activated, that is, by its squared norm $\|\textbf{b}_i\|^2$. A recent reference \cite{hyde2021obtaining} explores a coordinate-wise approach with a more thorough discussion of the coordinate order; they apply matching pursuit with pruning to estimate a sparse set of weights, yet without the advantage of a known structure of the blendshape rig function, leading to a computationally intensive method. 


\subsection{Contributions}

This paper follows a similar direction as \cite{seol2011artist} but addresses several main issues. In the first place, we target a more complex blendshape model used in modern production for highly-realistic human faces. Besides the linear terms in (\ref{eq:linear_blendshape_model}), corrective terms for some pairs or tuples of blendshapes are included \cite{seo2011compression, wu2018deep, rackovic2022majorization}. A corrective term for the pair of blendshapes $\textbf{b}_i$ and $\textbf{b}_j$ is denoted as $\textbf{b}^{\{ij\}}\in\mathbb{R}^{3n}$, and its activation weight is set to the product $w_iw_j$. These vectors are called first-level corrections, and in a similar manner, one can include corrections of higher levels for tuples of three or four blendshapes. In our experiments, we will assume blendshape models with three levels of corrections, and hence the blendshape function (here with a subscript $Q$ for \textit{quartic}) is given by
\begin{equation}\label{eq:q_blendshape_model}
\begin{split}
    f_Q(\textbf{w}) = &  \textbf{b}_0 + \textbf{Bw} + \sum_{\{i,j\}\in\mathcal{P}}w_iw_j\textbf{b}^{\{ij\}} +  \sum_{\{i,j,k\}\in\mathcal{T}}w_iw_jw_k\textbf{b}^{\{ijk\}} \\ 
    & + \sum_{\{i,j,k,l\}\in\mathcal{Q}}w_iw_jw_kw_l\textbf{b}^{\{ijkl\}}
\end{split}
\end{equation}
where $\mathcal{P},\mathcal{T},$ and $\mathcal{Q}$ are sets of pairs, triplets, and quadruplets of blendshapes that involve a corresponding corrective term, respectively. 

\paragraph{Fitting an arbitrarily accurate blendshape rig}
Compared with other methods that involve corrective terms \cite{rackovic2022majorization}, the approach proposed here leads to computationally efficient solutions under the corrections of any order. Our method is general enough to work with any level of corrections. To illustrate this, we will also include a case when a linear approximation of the rig is used in the experiments. This is in contrast with \cite{rackovic2022majorization} that can only handle first-order corrections. The reason why we restrict our experiments to the third level of corrective terms is that the animated models at our disposal do not have additional levels of corrections.

\paragraph{A solution with low cardinality}
Additionally, we add an $l_1$ norm regularizer to penalize the solution cardinality further. Finally, we also address the fact that \cite{seol2011artist} performs only a single pass over the weights. This can be seen as one step of a coordinate descent minimization algorithm. Even though a single pass of the algorithm gives a relatively good estimate of the solution, we will show in our experiments that adding several algorithm steps to our data fitting procedure can provide a better mesh fit while keeping the cardinality similar. 

The numerical results in Section \ref{sec:NumRes} show that our method outperforms the state-of-the-art approaches, giving an accurate mesh reconstruction with a mesh error lower or comparable to the best-performing methods and simultaneously reducing the cardinality of the weight vector over $20\%$. It further allows one to work with an arbitrary number of blendshape corrections, while the previous papers consider either a linear blendshape model or only a single correction level. Finally, the proposed method gives smooth temporal transitions, as shown for several animated sequences in the supplementary materials. This is also confirmed by the roughness metric (see Section \ref{sec:NumRes}), as our method shows more than double of a reduction compared to the benchmarks of a similar mesh error level.


\subsection{Notation}

Throughout this paper, scalar values will be denoted with lowercase Latin $a$, $b$, $c$, or lowercase Greek $\alpha,\beta,\gamma$ letters. Vectors are denoted with bold lowercase letters, e.g.,  $\textbf{a}$, $\textbf{b}$, $\textbf{c}$ and are indexed using a subscript, i.e., the $i^{th}$ element of a vector $\textbf{a}$ is $a_i$. If there is a subscript and the letter is still in bold, it is not indexing --- we will use this to distinguish blendshape vectors ($\textbf{b}_0,\textbf{b}_1,...,\textbf{b}_m$) as they have similar properties. We use $\textbf{0}$ and $\textbf{1}$ to denote vectors of all zeros and all ones, respectively. When we use order relations ($\geq,\leq,=$) between two vectors, they are assumed component-wise. Matrices are written in bold capital letters, e.g., $\textbf{A}$, $\textbf{B}$, $\textbf{C}$. Functions are given using lowercase letters, but with their arguments enclosed in parenthesis, e.g., $f(\cdot),g(\cdot)$. The Euclidean norm is denoted by $\|\cdot\|.$


\section{Related Work}

Blendshape animation is an attractive research topic because of its high relevance in our perception of the human face and the need for more realistic face representation. While anatomically-based face models might produce greater fidelity in realistic deformations and perception \cite{sifakis2005automatic, ichim2017phace}, they are usually much harder to animate and adjust manually and lack interpretability. Blendshape models have been studied in the literature as early as the end of the last century \cite{pinghin1998, choe2001analysis, choe2001performance}, and several main research directions have been established. The first challenge in the model is creating the blendshape basis since it can take from a few dozen up to several hundred blendshapes, so it is a time and labor-intensive task. Several papers propose automated solutions for producing the basis. In \cite{deng2006animating, buoaziz2013online}, the authors apply principal component analysis (PCA) over a dense motion capture; however, while PCA-based meshes are well suited for automated animation \cite{moser2021semi}, they lack explainability, making them undesirable to artists. A different approach studied by \cite{li2010example, li2013realtime, ribera2017facial} considers a pre-sculpted generic blendshape basis, that is used to create a semantically equivalent basis for a custom character applying a deformation transfer. Using a similar approach, \cite{chaudhuri2020personalized} trains a deep learning method to estimate the  person-specific blendshapes from video. Our paper does not provide contributions in this aspect and in the method we assume that a blendshape basis is given apriori, and that it closely resembles the actor/user.

The next challenge is adjusting controller weights to produce an animation. This step can be automated if there is a reference motion in the form of a 4D scan or motion capture of markers on the actor's face. This problem is called the inverse rig problem or automatic keyframe animation, and there are two main approaches to solving it: model-based and data-based methods. Model-based solutions rely on optimization techniques for model fitting and demand a precise definition of a rig function, while data is used for fitting model parameters. The problem is usually formalized as a least-squares problem, and regularization is often added to enhance the desired behavior, like stability \cite{ccetinaslan2016position, cetinaslan2020sketching, cetinaslan2020stabilized}, sparsity \cite{buoaziz2013online, neumann2013sparse, rackovic2022majorization}, or temporal smoothness \cite{tena2011interactive, seol2012spacetime}. On the other side, data-based solutions can work with an arbitrary rig function \cite{song2020accurate} but demand vast amounts of data for training a good predictor. Common models here are Radial Basis Function-based regressors \cite{Song2011CharacteristicFR, seol2014tuning}, Gaussian Processes Regression \cite{holden2015learning, reverdy2015optimal} and Neural Networks \cite{bailey2020fast, chaudhuri2020personalized}. As a final step, the meshes obtained as a solution to the inverse rig problem are combined with albedo maps \cite{feng2021learning}, lighting conditions, and material parameters to render the person-specific skin details and produce a life-like output image \cite{laine2020modular,lombardi2018deep, thies2018headon}. The problem of rig inversion is the main focus of our paper, and we propose a model-based approach that solves a constrained $l_1$ norm regularized non-linear least squares. The algorithm takes into account complex corrective blendshape terms, hence allowing high accuracy of the mesh fit, and yet, due to the coordinate descent approach and sparsity regularization, the obtained weight vectors have low cardinality making posterior manual adjustments possible.

Similar to the problem of the inverse rig is that of direct manipulation. However, it demands a real-time solution, since it assumes a deformation is propagated while the user is dragging the vertices of a character to adjust a given expression. To avoid artifacts that appear when all the non-selected markers in the face are kept fixed, \cite{lewis2010direct} proposes a general model where controllers are fitted taking into account an arbitrary number of selected manipulators. This idea is further developed in \cite{seo2011compression}, paying special attention to local influences of blendshapes. Later, \cite{cetinaslan2020sketching} develops a sketch-based method that leads to more intuitive manipulation, and the authors further improve the method in \cite{cetinaslan2020stabilized}.

Another direction of interest in blendshape animation is the segmentation of the face, and there is a number of approaches based on the final intention for the obtained segments. The main categories here are a localized or distributed approach to solving the inverse rig \cite{joshi2006learning, tena2011interactive, hirose2012creating, reverdy2015optimal, fratarcangeli2020fast, bailey2020fast, rackovic2021clustering}, where the mesh segments are in general relatively big, and adding secondary motion effects \cite{zoss2020data} to increase the plausibility of already animated characters \cite{neumann2013sparse, wu2016anatomically, romeo2020data}, where, in general, one produces a large number of very small segments.

This paper focuses on a model-based approach to solving the inverse rig problem. The animated avatars are highly realistic pre-sculpted blendshape models with additional corrections levels. Reference frames are given in the form of 3D face scans, and an imperative is on the high-accuracy expression reconstruction.
 

\section{Proposed Method}\label{sec:coord_desc}

This section introduces our algorithm for an accurate solution to the inverse rig problem for complex blendshape models. It is desirable for a solution vector $\textbf{w}$ to be sparse while producing an accurate reconstruction of a given mesh $\widehat{\textbf{b}}$. Additionally, weights must stay within a $[0,1]$ interval to respect the construction of animated characters. We pose the optimization problem as 
\begin{equation}\label{eq:q_problem}
    \minimize_{\textbf{0}\leq\textbf{w}\leq\textbf{1}} \frac{1}{2} \| f(\textbf{w}) - \widehat{\textbf{b}} \|^2 + \alpha \textbf{1}^T\textbf{w},
\end{equation}
with $\alpha \geq 0$ being a regularization parameter. Note that, due to non-negativity constraints on $\textbf{w}$, the regularization term is equal to the $l_1$ norm, which is known as a sparsity-enhancing regularizer. A rig function $f(\textbf{w})$ is given without a subscript because we want a solution that would work with arbitrary complexity of the rig, for example, linear or quartic. As the animated characters at our disposal consist of up to three levels of correction, the highest accuracy is achieved with a quartic rig defined in (\ref{eq:q_blendshape_model}), hence we give a derivation (and experimental results) for $f(\textbf{w}):=f_Q(\textbf{w})$. Nevertheless, we also include the case when $f(\textbf{w}):=f_L(\textbf{w})$, since the linear blendshape model is the one used often in the literature. We will refer to the two approaches as $\textit{Quartic}$ and $\textit{Linear}$, respectively. 

For the \textit{Quartic} case, a non-linearity of the rig makes the problem (\ref{eq:q_problem}) non-convex and hard to solve if weights \textbf{w} are to be estimated jointly. However, we can approach the problem component-wise, and then have a sequence of quadratic programs instead. That is, for a controller $i\in\{1,...,m\}$ we assume that all the weights $w_j$, for $j\neq i$, are fixed (to initial or previously estimated value), while we only need to estimate $w_i$, by solving
\begin{equation}\label{eq:q_component_problem}
    w_i \leftarrow \argmin_{0\leq w\leq 1} \frac{1}{2} \| w\boldsymbol\phi_i + \boldsymbol\psi_i \|^2 + \alpha w.
\end{equation}
Here $\boldsymbol\phi_i\in\mathbb{R}^{3n}$ is a vector that contains all the blendshape components (i.e., blendshape vector and corrective terms) that participate in the product with $w_i$:
\begin{equation}\label{eq:phi}
    \begin{split}
       \boldsymbol\phi_i & = \textbf{b}_i + \sum_{\{i,j\}\in\mathcal{P}}w_j\textbf{b}^{\{ij\}} +  
       \sum_{\{i,j,k\}\in\mathcal{T}}w_jw_k\textbf{b}^{\{ijk\}} + \\
       & \sum_{\{i,j,k,l\}\in\mathcal{Q}}w_jw_kw_l\textbf{b}^{\{ijkl\}};
    \end{split}
\end{equation}
and $\boldsymbol\psi_i\in\mathbb{R}^{3n}$ contains all the other components, together with a given target mesh $\widehat{\textbf{b}}$:
\begin{equation}\label{eq:psi}
    \begin{split}
       \boldsymbol\psi_i & = \sum_{j\neq i}w_j\textbf{b}_j + \sum_{\substack{\{j,k\}\in\mathcal{P}\\j,k\neq i}}w_jw_k\textbf{b}^{\{jk\}} + \sum_{\substack{\{j,k,l\}\in\mathcal{T}\\j,k,l\neq i}}w_jw_kw_l\textbf{b}^{\{jkl\}} + \\
       & \sum_{\substack{\{j,k,l,h\}\in\mathcal{Q}\\j,k,l,h\neq i}}w_jw_kw_lw_h\textbf{b}^{\{jklh\}} - \widehat{\textbf{b}}.
    \end{split}
\end{equation}
The global solution to (\ref{eq:q_component_problem}) is found in closed form, by setting the derivative with respect to $w_i$ to zero, and projecting to the feasible set:
\begin{equation}\label{eq:quartic_solution}
     w_i \leftarrow P_{[0,1]}\left(\frac{\boldsymbol\phi_i^T\boldsymbol\psi_i - \alpha}{\|\boldsymbol\phi_i\|^2}\right),
\end{equation}
where the projection operator is defined as
\begin{equation}
P_{[0,1]}(x) = 
    \begin{cases}
       0, & \text{if } x<0, \\
       1, & \text{if } x>1, \\
       x, & \text{otherwise}.
    \end{cases}
\end{equation}

For the \textit{Linear} case, a coordinate optimization problem analogue to (\ref{eq:q_component_problem}) is
\begin{equation}\label{eq:l_component_problem}
    w_i \leftarrow \argmin_{0\leq w\leq 1} \frac{1}{2} \| w\textbf{b}_i + \sum_{j\neq i}w_j\textbf{b}_j - \widehat{\textbf{b}} \|^2 + \alpha w,
\end{equation}
with the solution 
\begin{equation}\label{eq:linear_solution}
     w_i \leftarrow P_{[0,1]}\left(\frac{\textbf{b}_i^T(\sum_{j\neq i}w_j\textbf{b}_j - \widehat{\textbf{b}}) - \alpha}{\|\textbf{b}_i\|^2}\right).
\end{equation}
Notice that (\ref{eq:quartic_solution}) and (\ref{eq:linear_solution}) have the same structure and that by setting the corrective blendshape terms of (\ref{eq:phi}) and (\ref{eq:psi}) to zero, solution (\ref{eq:quartic_solution}) simplifies to (\ref{eq:linear_solution}). This confirms that our approach is general enough to work with an arbitrary number of correction levels.

When the optimal component-wise weights are estimated for all $m$ controllers, the process is repeated until some stopping criterion is satisfied. This type of iteration is known as coordinate descent, and it is guaranteed to produce monotonically non-increasing costs \cite{luo1992convergence, wright2015coordinate}. In the case when a model is defined via a linear rig function (\ref{eq:linear_blendshape_model}), the objective (\ref{eq:q_problem}) consists of a smooth convex function $\|\textbf{Bw}-\widehat{\textbf{b}} \|^2$ and a convex and separable regularization term $\textbf{1}^T\textbf{w}$, hence we can claim that our method converges to the optimal solution of (\ref{eq:q_problem}), as proved in \cite{wright2015coordinate}.
The pseudo-code of the proposed method is given in Algorithm \ref{alg:pseudocode}. 

\begin{algorithm}
\caption{}\label{alg:pseudocode}
\begin{algorithmic}
\Require A mesh vector $\widehat{\textbf{b}}\in\mathbb{R}^{3n}$, a set of blendshape vectors $\textbf{b}_1,...,\textbf{b}_m\in\mathbb{R}^{3n}$ ordered as $\|\textbf{b}_1\|^2 \geq \|\textbf{b}_2\|^2 \geq \cdots \geq \|\textbf{b}_m\|^2$ (\textit{we implicitly order blendshape vectors by norm magnitude without loss of generality}), regularization parameter $\alpha\geq 0$, and the number of iterations $T\in\mathbb{N}$. If a considered rig function is quartic, include also corrective terms $\textbf{b}^{\{ij\}}$ for $\{i,j\}\in\mathcal{P}$, $\textbf{b}^{\{ijk\}}$ for $\{i,j,k\}\in\mathcal{T}$, $\textbf{b}^{\{ijkl\}}$ for $\{i,j,k,l\}\in\mathcal{Q}$.
\Ensure Optimal weight vector $\textbf{w}\in\mathbb{R}^m$ such that $\textbf{0}\leq\textbf{w}\leq\textbf{1}$.
\State Initialize weight vector by $\textbf{w}=\textbf{0}$
\For{t=1,...,T}
    \For{i=1,...,m}
        \If{Rig function is linear}
            \State $w_i \leftarrow P_{[0,1]}\left(\frac{\textbf{b}_i^T(\sum_{j\neq i}w_j\textbf{b}_j - \widehat{\textbf{b}}) - \alpha}{\|\textbf{b}_i\|^2}\right)$
        \ElsIf{Rig function is quartic}
            \State Compute $\boldsymbol\phi_i$ from (\ref{eq:phi}) and $\boldsymbol\psi_i$ from (\ref{eq:psi})
            \State $w_i \leftarrow P_{[0,1]}\left(\frac{\boldsymbol\phi_i^T\boldsymbol\psi_i - \alpha}{\|\boldsymbol\phi_i\|^2}\right)$
        \EndIf
    \EndFor
\EndFor
\end{algorithmic}
\end{algorithm}

Notice that our approach has some similarities with \textit{Seol}. If one sets $\alpha=0$ and removes the upper constraint $w\leq 1$, equation (\ref{eq:linear_solution}) becomes equivalent to the \textit{Seol} update rule from (\ref{eq:seol_iteration}). For this reason, we adopt some of the tactics from \cite{seol2011artist}. 

Any feasible vector $\textbf{w}\in\mathbb{R}^m$ can be used for the initialization; however, we stress that by initializing with a non-zero weight vector, the method cannot guarantee that mutually exclusive controllers will not get activated simultaneously, as explained in \cite{seol2011artist}. 

In a sequential mesh fitting, the order in which blendshapes are visited plays an important role, so we adapt the strategy from \cite{seol2011artist} to order them by the overall displacement magnitude:
\begin{equation}\label{eq:ordering}
    \|\textbf{b}_1\|^2 \geq \|\textbf{b}_2\|^2 \geq \cdots \geq \|\textbf{b}_m\|^2.
\end{equation}
This choice is inspired by the intuition behind the manual process, where an artist tends to first set the weights of the controllers of more drastic motions (like mouth opening or nose squeezing) before visiting more subtle ones. This has some similarities with a sparse solution of the matching pursuit problem \cite{mallat1993matching}, yet we will discuss alternative strategies for choosing the optimization order in Section \ref{sec:order}.

One of the relevant points not discussed in \cite{seol2011artist} was the possibility of multiple algorithm passes --- the authors terminated the process after each controller was visited and updated once. We conclude from the theory on coordinate descent and from our experiments that increasing the number of passes leads to a significant reduction in mesh error at the cost of a slight increase of the solution cardinality. 

In the next section, we show the performance of our algorithm on a set of animated human characters, and benchmark against state-of-the-art methods for solving the inverse rig problem. We also give an extensive results analysis and discuss the main improvements compared to the baseline \textit{Seol} approach and alternative strategies for some aspects of the algorithm.


\section{Evaluation}\label{sec:results}

This section compares the solutions of different approaches on several data sets and gives an extensive discussion of the results.

\subsection{Data}

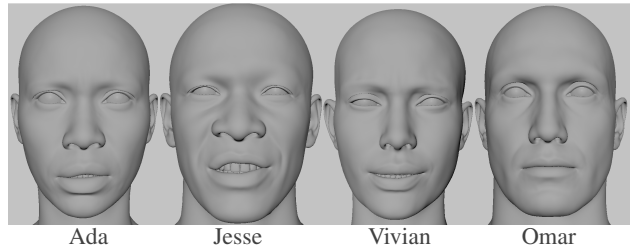
\begin{figure}
    \centering
    \begin{tikzpicture}
    \node[above right, inner sep=0] (image) at (0,0){\includegraphics[width=0.5\linewidth]{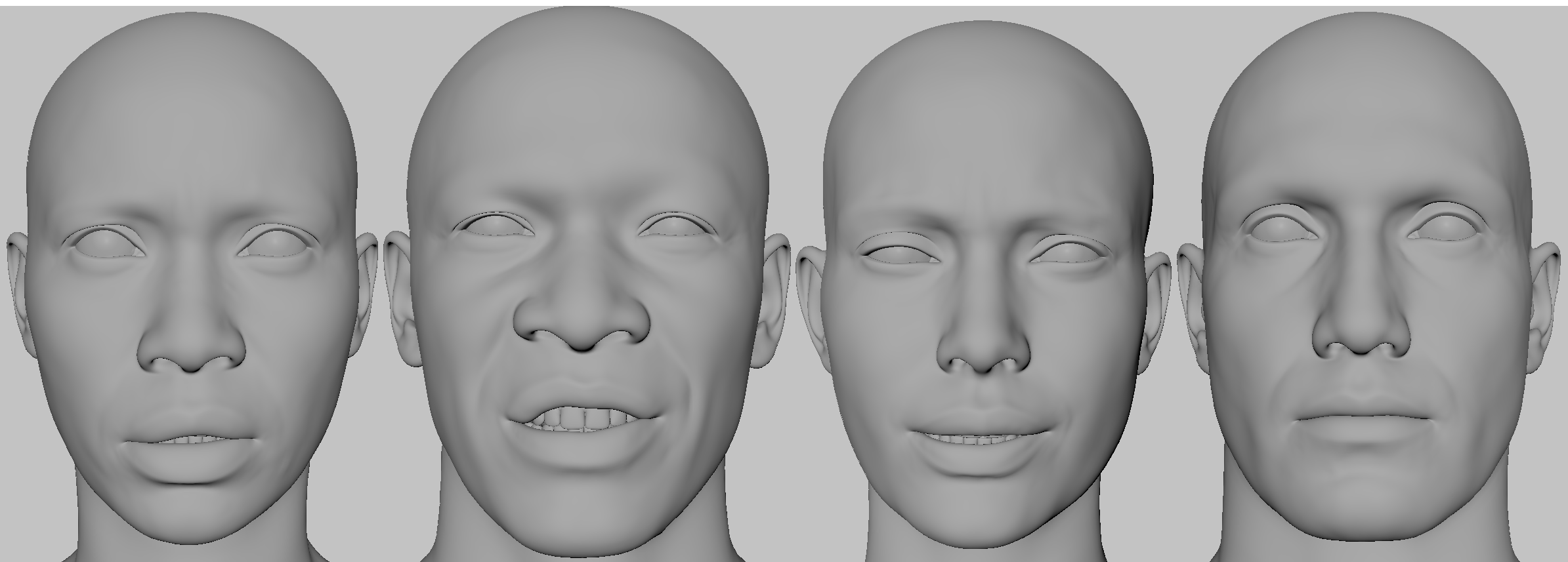}};
    \begin{scope}[
        x={($0.1*(image.south east)$)},
        y={($0.1*(image.north west)$)}]
      \node[darkgray] at (1.30,-0.5) {\small Ada };
      \node[darkgray] at (3.70,-0.5) {\small Jesse };
      \node[darkgray] at (6.30,-0.5) {\small Vivian };
      \node[darkgray] at (8.70,-0.5) {\small Omar };
    \end{scope}
    \end{tikzpicture}
    \caption{Head models available at MetaHuman creator.}
    \label{fig:metahumans}
\end{figure}

 Experiments are performed over five animated avatars, four of which are freely available on the \texttt{MetaHuman Creator}\footnote{\url{https://www.unrealengine.com/en-US/metahuman}} --- \textit{Ada, Jesse, Vivian} and \textit{Omar}, and shown in Figure \ref{fig:metahumans}. The fifth character is a proprietary model provided by 3lateral studio \footnote{\url{https://www.3lateral.com}}, for the purpose of this research, and we refer to it as \textit{Char 5}. The size of the head is comparable over the characters, with a width (distance from left to right ear) of around $18cm$. However, the number of vertices and controllers differ, and we give it summarized in Table \ref{tab:stat1}.

\begin{table}[]
    \centering
    \begin{tabular}{c | c c c c c}
                    & Ada     & Jesse    & Vivian  & Omar & Char 5  \\
                  \hline
                $m$ & $102$   & $102$    & $102$   & $130$    & $147$ \\
                $n$ & $10000$ & $10000$  & $10000$ & $3746$   & $2511$\\
                $N$ & $600$   & $600$    & $600$   & $600$    & $600$ \\
    $|\mathcal{P}|$ & $185$   & $185$    & $185$   & $187$    & $160$ \\
    $|\mathcal{T}|$ & $130$   & $130$    & $130$   & $130$    & $68$ \\
    $|\mathcal{Q}|$ & $50$    & $50$     & $50$    & $50$     & $12$ \\
    \end{tabular}
    \caption{Dimensions of the four animated characters, where $m$ is the number of blendshapes in the basis, $n$ is the number of vertices in face mesh, $N$ is the number of test frames, and $|\mathcal{P}|$, $|\mathcal{T}|$ and $|\mathcal{Q}|$ are the numbers of corrective combinations of first, second and third order, respectively.}
    \label{tab:stat1}
\end{table}

Although it would be ideal to perform experiments on the actual 3D facial scans matching the avatars, such high-fidelity data is, in general, costly and only produced for profit applications, and it is not available for researchers. For this reason, we will have to restrict ourselves to working with synthetic data, yet we will include noise in our fitting data in order to mimic the meshes acquired with a 3D laser scanner or photogrammetry \cite{vasiljevic2021copyright, cui20103d}. In Figure \ref{fig:adaNoise}, we show a close shoot of \textit{Ada} with varying amounts of added Gaussian noise \cite{wand2007reconstruction, sun2008noise}. The top left subfigure is a clean mesh, followed by the meshes with the increasing variance of the added noise. We notice that with low values, like $\sigma^2=0.01$, the mesh is almost indistinguishable from the clean one, while one order higher value gives a mesh that is too corrupted. We chose to work with $\sigma^2=0.03$, as it produces a smoothness effect similar to that obtained by modern 3D scanners, yet later in this section, we will also see how varying noise levels affect the results of different methods. This noise is only added on the target meshes $\widehat{\textbf{b}}$ in the process of fitting, i.e., solving the inverse rig, while the error for the reconstructed meshes is computed with respect to the original, noise-free data.

\begin{figure}
    \centering
    \begin{tikzpicture}
    \node[above right, inner sep=0] (image) at (0,0){\includegraphics[width=0.5\linewidth]{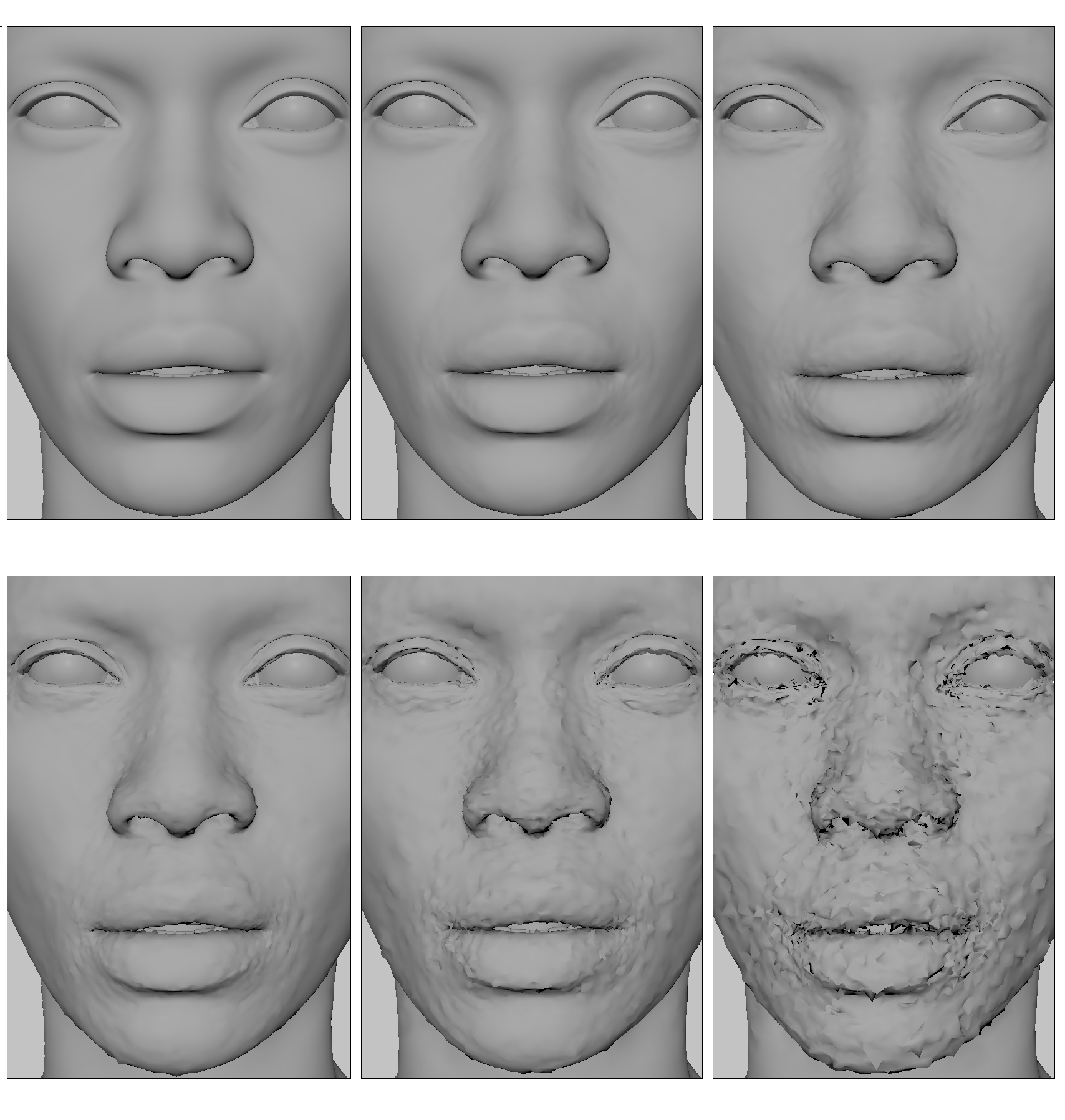}};
    \begin{scope}[
        x={($0.1*(image.south east)$)},
        y={($0.1*(image.north west)$)}]
      \node[darkgray] at (1.80,5.1) {\small $0.00$ };
      \node[darkgray] at (5.00,5.1) {\small $0.01$ };
      \node[darkgray] at (8.30,5.1) {\small $0.02$ };
      \node[darkgray] at (1.80,0.1) {\small $0.03$ };
      \node[darkgray] at (5.00,0.1) {\small $0.05$ };
      \node[darkgray] at (8.30,0.1) {\small $0.10$ };
    \end{scope}
    \end{tikzpicture}
    \caption{Face mesh of \textit{Ada} corrupted with different noise levels. The upper left is the original clean mesh, and each other represents a mesh with added Gaussian noise with the standard deviation $\sigma^2$ corresponding to the value below the figure.}
    \label{fig:corrupted}
\end{figure}
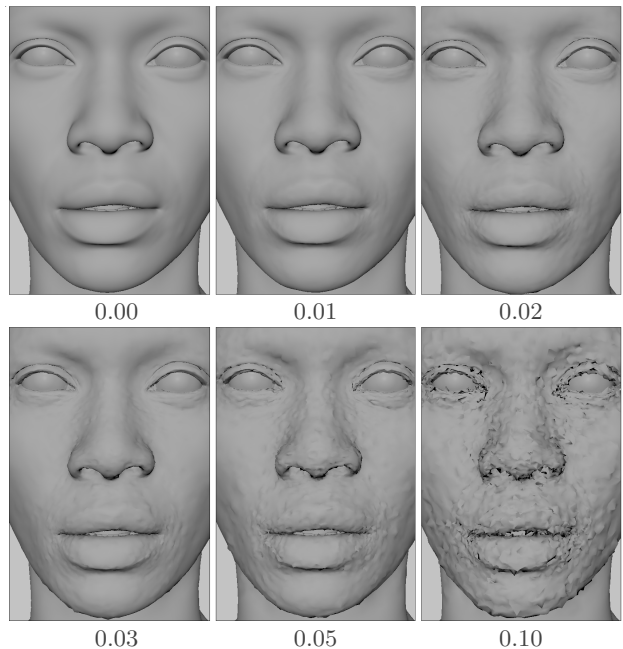

In addition to Figure \ref{fig:corrupted}, Figure \ref{fig:adaNoise} compares the results over training frames for \textit{Ada} under different noise levels, as indicated by the color of the dots, for the proposed method and for two benchmark methods, \textit{Seol} and \textit{Çetinaslan}. (See Section \ref{sec:NumRes} for details on benchmark methods, and Section \ref{sed:metrics} for evaluation metrics.) The results are shown for each of 100 data points, and additionally, since \textit{Quartic} and \textit{Çetinaslan} demand a value of the regularization parameter $\alpha$, we include choices of $\alpha\in\{0,0.1,1\}$, as indicated by the size and opacity of the dots. One can notice that \textit{Çetinaslan} is more affected by the increase of the noise levels --- for low values of $\sigma^2$, it gives the lowest mesh error, while with the increase, the error is drastically higher. On the other side, the proposed method is quite robust in this sense, and additionally, with an increased value of regularizer parameter $\alpha$, it leads to a significant reduction of the cardinality, without visibly affecting the mesh error. 

\begin{figure}
    \centering
    \includegraphics[width=0.5\linewidth]{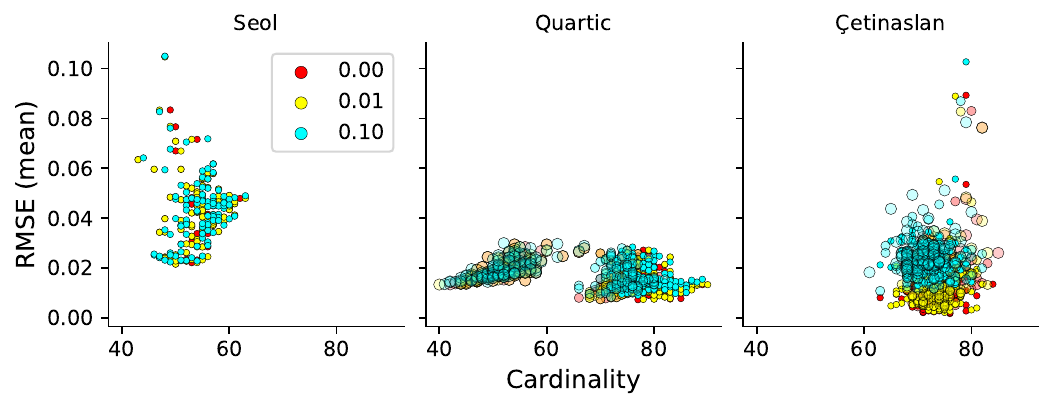}
    \caption{Point-wise predictions for three methods (\textit{Seol} as proposed in \cite{seol2011artist}, \textit{Quartic} proposed in this paper, and \textit{Çetinaslan} from \cite{cetinaslan2020sketching}) with respect to different noise levels. Dot colors correspond to the standard deviation $\sigma^2 \in \{0,0.01,0.1\}$ of the added noise, while the dot sizes in \textit{Quartic} and \textit{Çetinaslan} are proportional to the regularization value $\alpha\in\{0,0.1,1\}$.}
    \label{fig:adaNoise}
\end{figure}


\subsection{Metrics}\label{sed:metrics}

The mesh error is an important metric of interest in a realistic face reconstruction. To measure this, we use the root mean squared error (RMSE) given by
\begin{equation}
    \text{RMSE}(\textbf{w}) =\sqrt{\frac{\|f_Q(\textbf{w})-\tilde{\textbf{b}}\|^2}{n}},
\end{equation}
where $f_Q(\textbf{w})$ is the reconstructed mesh vector and $\tilde{\textbf{b}}$ is a noise-free target mesh. We remind the reader that as an input to the algorithms in our experiments, we used meshes corrupted by Gaussian noise, $\widehat{\textbf{b}} = \tilde{\textbf{b}} + \epsilon$, such that $\epsilon \sim \mathcal{N}(0,\sigma^2),$ and evaluated the results on the clean meshes $\tilde{\textbf{b}}$. Recall that we work with meshes with thousands of vertices, hence the average of the error over all of them might not be able to indicate if the obtained expression resembles the original well or not. It is important whether the error is visible, i.e., perceivable by a human, in a reconstructed face, hence it bears more significance if a small number of vertices gives a large offset than if each vertex in the face is only slightly off. To account for this, besides the mean over all the $n$ vertices, we will also show the $95^{th}$ percentile of the error over the mesh vertices. Besides the mesh error, we are also interested in the cardinality of the solution, i.e., the number of non-zero weights, since the frames with a large number of activated weights might be unstable and hard to adjust by an artist. We will complement this with the $l_1$ norm as well since sometimes it is also used for approximately measuring the sparsity of the solution. Finally, since in the animation temporal continuity or smoothness is an important concept, we include another metric that measures the temporal roughness of the weight curves. The roughness is inversely proportional to the smoothness, and higher values of this metric indicate that the results are less smooth. The metric is based on second-order differences \cite{marquis2022rig}
\begin{equation}
    \text{Roughness}(w_i) = \sum_{t=2}^{T-1} \left(w_i^{(t-1)} - 2w_i^{(t)} + w_i^{(t+1)}\right)^2,
\end{equation}
where the score of zero is obtained for a constant vector and increases the more the consecutive entries differ. Notice that, while the other metrics introduced so far are computed for a single frame (and over the weights or vertices), \textit{Roughness} is computed for a single weight over the frames.
 
\begin{figure}
    \centering
    \includegraphics[width=0.5\linewidth]{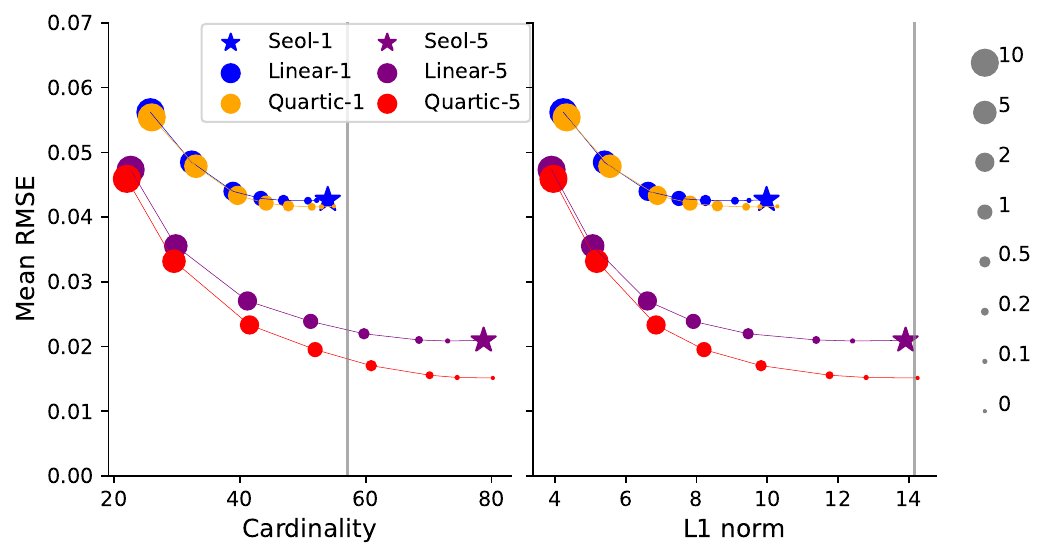}
    \caption{Trade-off between the mesh error and cardinality / L1 norm of the weight vector. A blue star (\textit{Seol-1}) is the solution proposed by \cite{seol2011artist}. Adding an $l_1$ regularization term to \textit{Seol-1} leads to the results presented as blue dots (\textit{Linear-1}), where dot sizes correspond to a value of regularization parameter $\alpha$. Further adjustment is adding multiple iterations of the algorithm (5 in this case), which is shown in purple, where again a star represents a non-regularized case (\textit{Seol-5}) and dot sizes correspond to the value of $\alpha$ (\textit{Linear-5}). Analog to \textit{Linear-1} and \textit{Linear-5} are \textit{Quartic-1} (orange) and \textit{Quartic-5} (red) respectively, where the linear rig function (\ref{eq:linear_blendshape_model}) is substituted by a quartic (\ref{eq:q_blendshape_model}). Gray vertical lines represent the values of the ground-truth data.}
    \label{fig:ttradeoff1}
\end{figure}


\subsection{Numerical Results}\label{sec:NumRes}

All the characters in our experiments have corrective terms and a rig function in the form (\ref{eq:q_blendshape_model}), hence it is reasonable to apply our method under the quartic rig function. Nevertheless, we will also include the case when a linear rig approximation is used, to show that our method may outperform the others even in the simplified case, denoting these two approaches \textit{Quartic} and \textit{Linear} respectively. As a first benchmark method, we will use \textit{Seol}, introduced earlier in Section \ref{sec:intro}. Since there are similarities between the proposed method and \textit{Seol}, let us first examine how the results of these two relate. These results are shown in Figure \ref{fig:ttradeoff1}. Our method with a linear rig approximation and a single iteration over the weights is denoted with \textit{Linear-1} (similarly, a linear model with 5 iterations is denoted \textit{Linear-5}, and in the same analogy we have \textit{Quartic-1} and \textit{Quartic-5}). In the case where the regularization term is set to $\alpha=0$, it simplifies to \textit{Seol}, as indicated by a blue star, while for the higher values of $\alpha\in\{0.1,0.2,0.5,1,2,5,10\}$, results are presented in blue dots of the corresponding sizes. We see that for low to medium values of $\alpha$, our method gives a visible reduction in cardinality (and $l_1$ norm) compared to \textit{Seol}, without affecting the mesh error, while for the higher values, a trade-off is made between accuracy and sparsity. In this case of a single iteration, the inclusion of corrective terms does not seem to have significant effects, as the curve of \textit{Quartic-1} closely follows \textit{Linear-1}. On the other side, with the increased number of iterations, we see that \textit{Quartic-5} gives a lower mesh error than \textit{Linear-5} for the same cardinality value. However, both these curves are under those of a single iteration, showing a large margin of improvement with respect to both axes. A purple star denoted \textit{Seol-5} indicates a modification of the solution from \cite{seol2011artist}, where the method is iterating through weights five times (as opposed to \textit{Seol-1}, which is the solution as proposed in \cite{seol2011artist}). \textit{Seol-5} exhibits more precise mesh fit than \textit{Seol-1}, yet at the cost of considerable increase in cardinality. 

\begin{figure}
    \centering
    \includegraphics[width=0.5\linewidth]{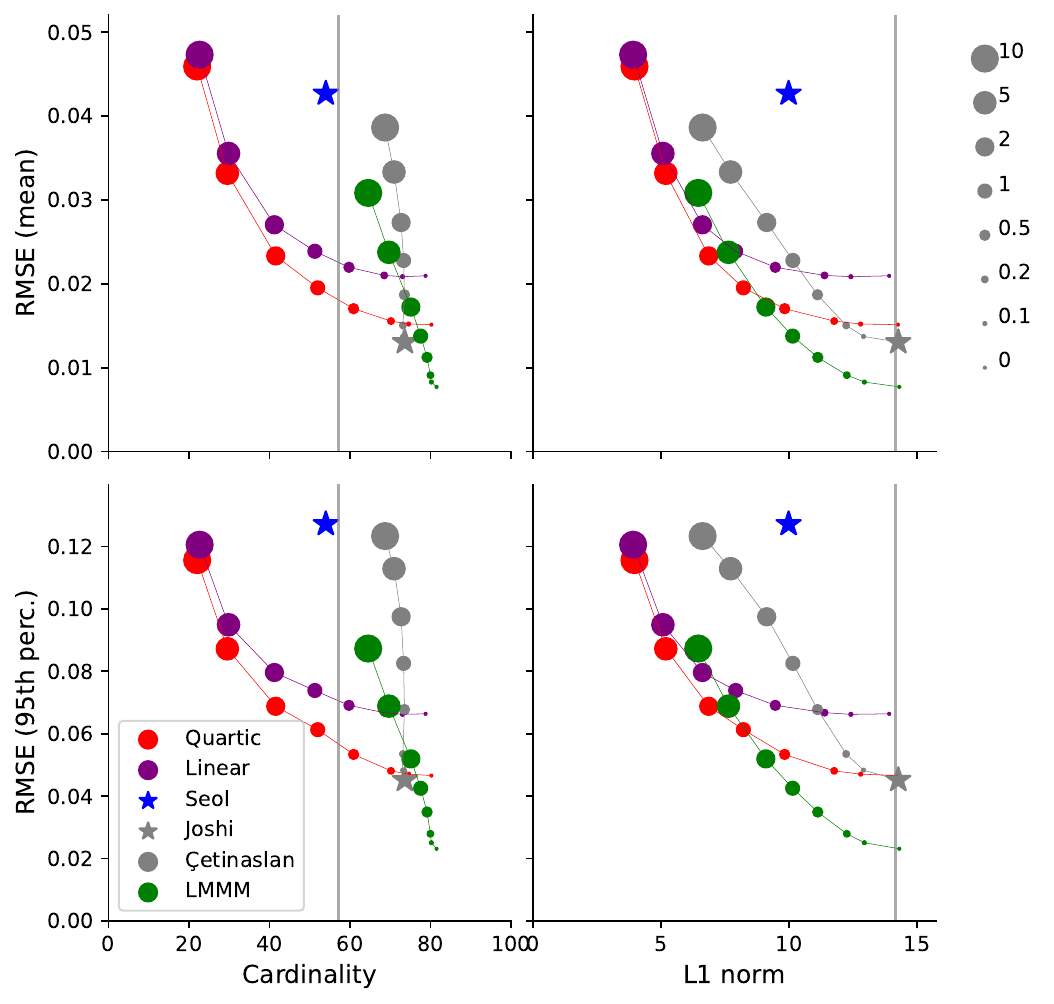}
    \caption{Trade-off between the mesh error and cardinality / L1 norm of the weight vector for \textit{Ada}. The top row shows the mean mesh error, while the bottom corresponds to the $95^{th}$ percentile of the error. Dot sizes are proportional to the size of the regularization parameter $\alpha$ ranging from $0$ to $10$, as indicated on the right. The proposed solution with the quartic rig function is shown in red (\textit{Quartic}), and the one with a linear approximation is in blue (\textit{Linear}). The solution proposed by \cite{cetinaslan2020sketching} is represented by gray dots (\textit{Çetinaslan}), where a star (\textit{Joshi}) corresponds to a special case whit no regularization \cite{joshi2006learning}. The approach of \cite{rackovic2022majorization} is shown in green (\textit{LMMM}). Gray vertical lines represent the values of the ground-truth data.}
    \label{fig:adaTraining}
\end{figure}

As opposed to these sequential approaches, in Section \ref{sec:intro}, we also mentioned a solution proposed by \cite{joshi2006learning} (\textit{Joshi}), that solves for all the blendshape weights jointly. While the approach of \textit{Joshi} is simple and often satisfactory for general purposes, we will consider two other methods from the same group that are more recent and might provide a better solution. 
The first one we refer to as \textit{Çetinaslan}, was proposed in \cite{cetinaslan2020sketching} as a generalization of the \textit{Joshi} solution. The method includes an $l_2$ squared regularization to problem (\ref{eq:psd_problem}), so a solution is obtained by solving a set of linear equations 
\begin{equation}
    (\textbf{B}^T\textbf{B} + 2\alpha \textbf{I})\textbf{w} = \textbf{B}^T\widehat{\textbf{b}}.
\end{equation}
The weights are afterward clipped to $[0,1]$ interval in order to satisfy the constraints. The other approach was proposed in \cite{rackovic2022majorization}, and to the best of our knowledge, it is the only model-based approach that includes corrective terms in the blendshape model when solving the inverse rig. It is, however, restricted to working with only the first-level corrections (as opposed to our proposed algorithm, which can take any number of corrective levels). The method considers the objective function
\begin{equation}
        \minimize_{\textbf{0}\leq\textbf{w}\leq\textbf{1}} \| \textbf{Bw} + \sum_{\{i,j\}\in\mathcal{P}}w_iw_j\textbf{b}^{\{ij\}} - \widehat{\textbf{b}} \|^2 + \alpha \textbf{1}^T\textbf{w},
\end{equation}
and applies Levenberg-Marquard \cite{Levenberg1944AMF, marquardt1963algorithm} and Majorization-Minimization \cite{hunter2004tutorial} to solve the problem iteratively, hence we refer to this approach as \textit{LMMM}. 

\begin{figure}
    \centering
    \includegraphics[width=0.5\linewidth]{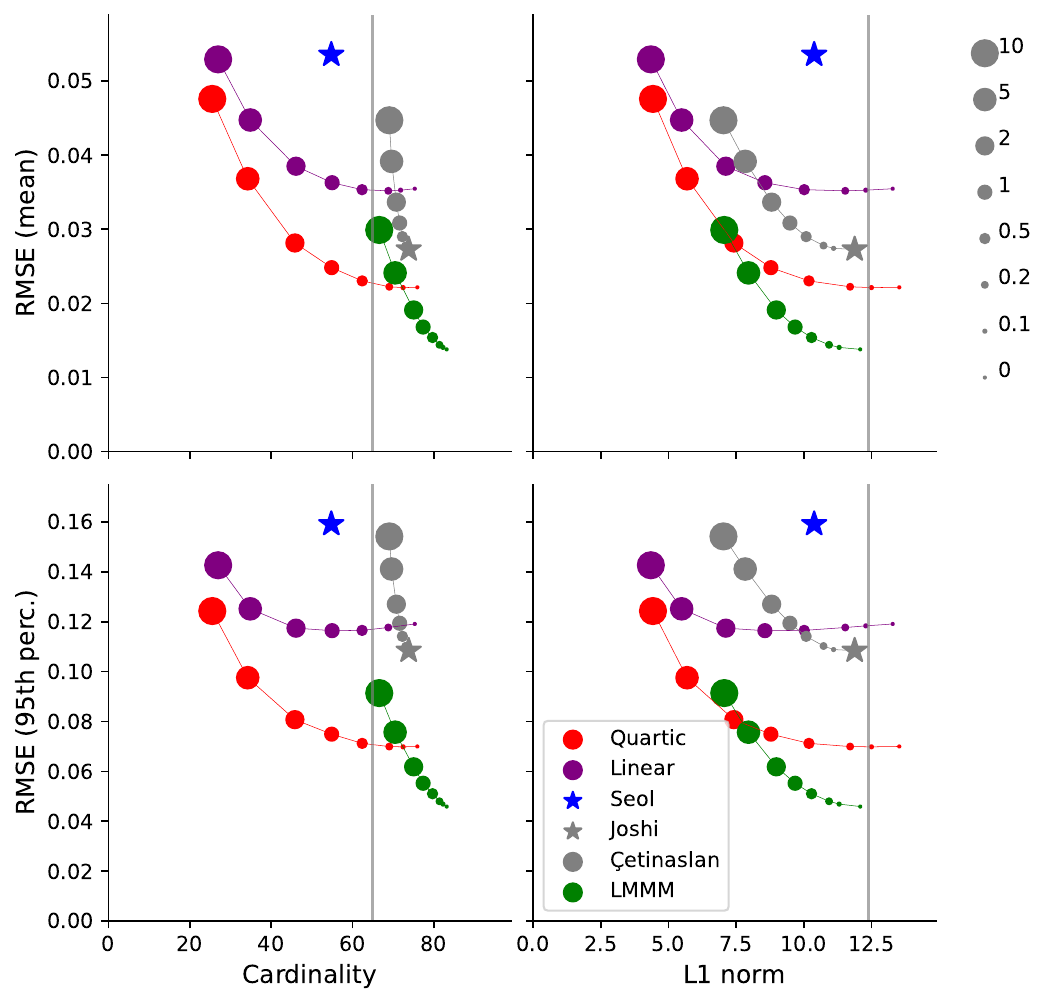}
    \caption{Trade-off between the mesh error and cardinality / L1 norm of the weight vector for \textit{Jesse}. The top row shows the mean mesh error, while the bottom corresponds to the $95^{th}$ percentile of the error. Dot sizes are proportional to the size of regularization parameter $\alpha$ ranging from $0$ to $10$, as indicated on the right.}
    \label{fig:jesseTraining}
\end{figure}

The first three benchmark approaches listed above, \textit{Seol}, \textit{Joshi} and \textit{Çetinaslan}, assume a linear rig function (\ref{eq:linear_blendshape_model}) when estimating the weights, while \textit{LMMM} assumes quadratic. Our method is tested with both linear and quartic rig (\textit{Linear} and \textit{Quartic} respectively). Nevertheless, once the weights are estimated, we use the full quartic rig to reconstruct the meshes with all the methods so that the evaluation results are fair.

\begin{figure}
    \centering
    \includegraphics[width=0.5\linewidth]{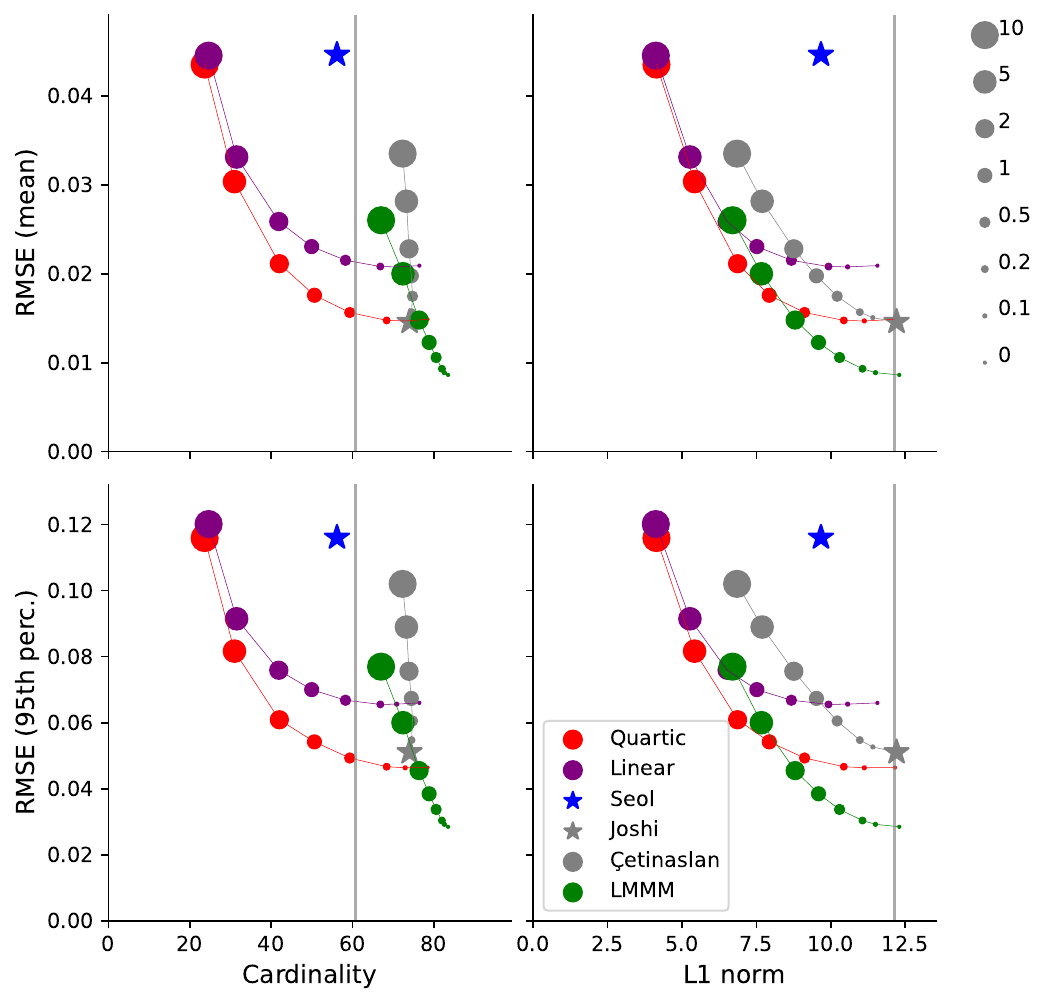}
    \caption{Trade-off between the mesh error and cardinality / L1 norm of the weight vector for \textit{Vivian}. The top row shows the mean mesh error, while the bottom corresponds to the $95^{th}$ percentile of the error. Dot sizes are proportional to the size of regularization parameter $\alpha$ ranging from $0$ to $10$, as indicated on the right.}
    \label{fig:vivianTraining}
\end{figure}

\begin{figure}
    \centering
    \includegraphics[width=0.5\linewidth]{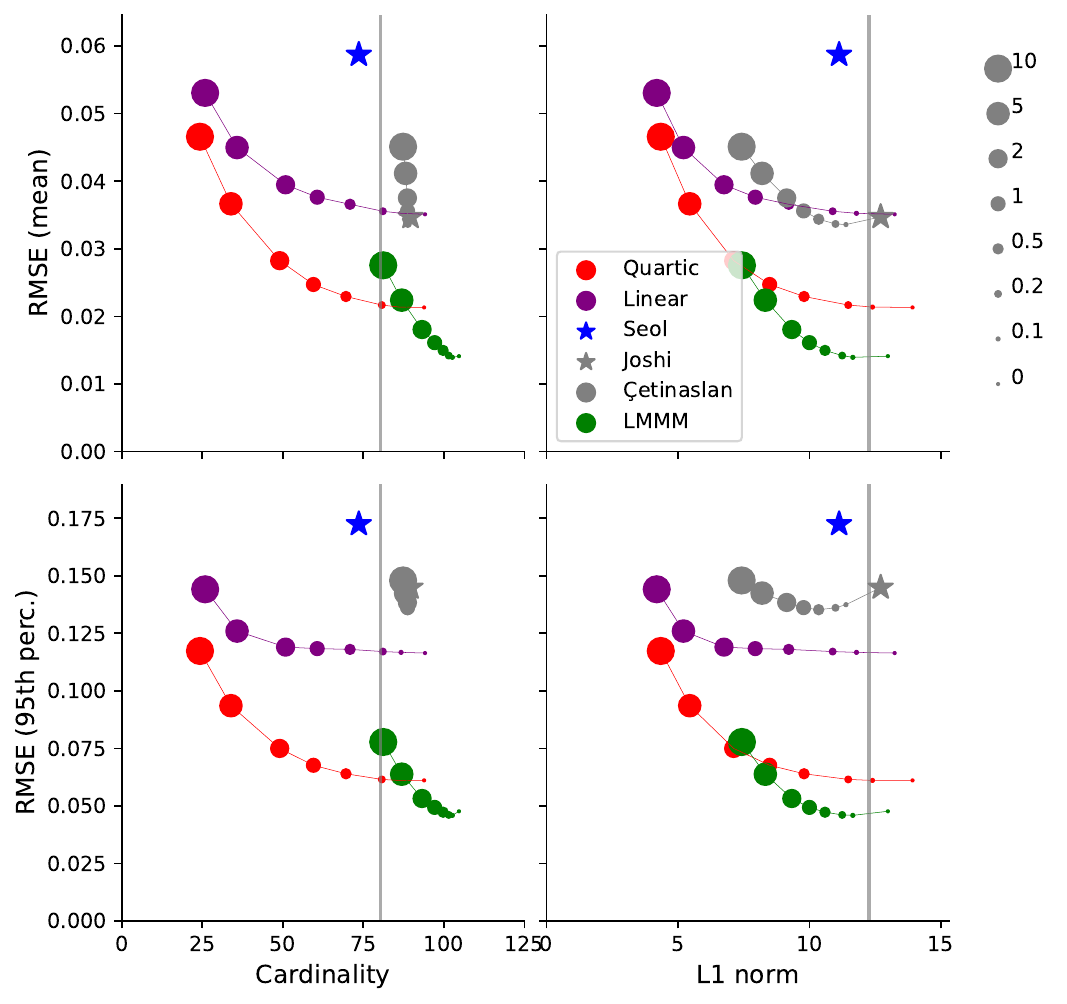}
    \caption{Trade-off between the mesh error and cardinality / L1 norm of the weight vector for \textit{Omar}. The top row shows the mean mesh error, while the bottom corresponds to the $95^{th}$ percentile of the error. Dot sizes are proportional to the size of regularization parameter $\alpha$ ranging from $0$ to $10$, as indicated on the right.}
    \label{fig:omarTrainig}
\end{figure}

\begin{figure}
    \centering
    \includegraphics[width=0.5\linewidth]{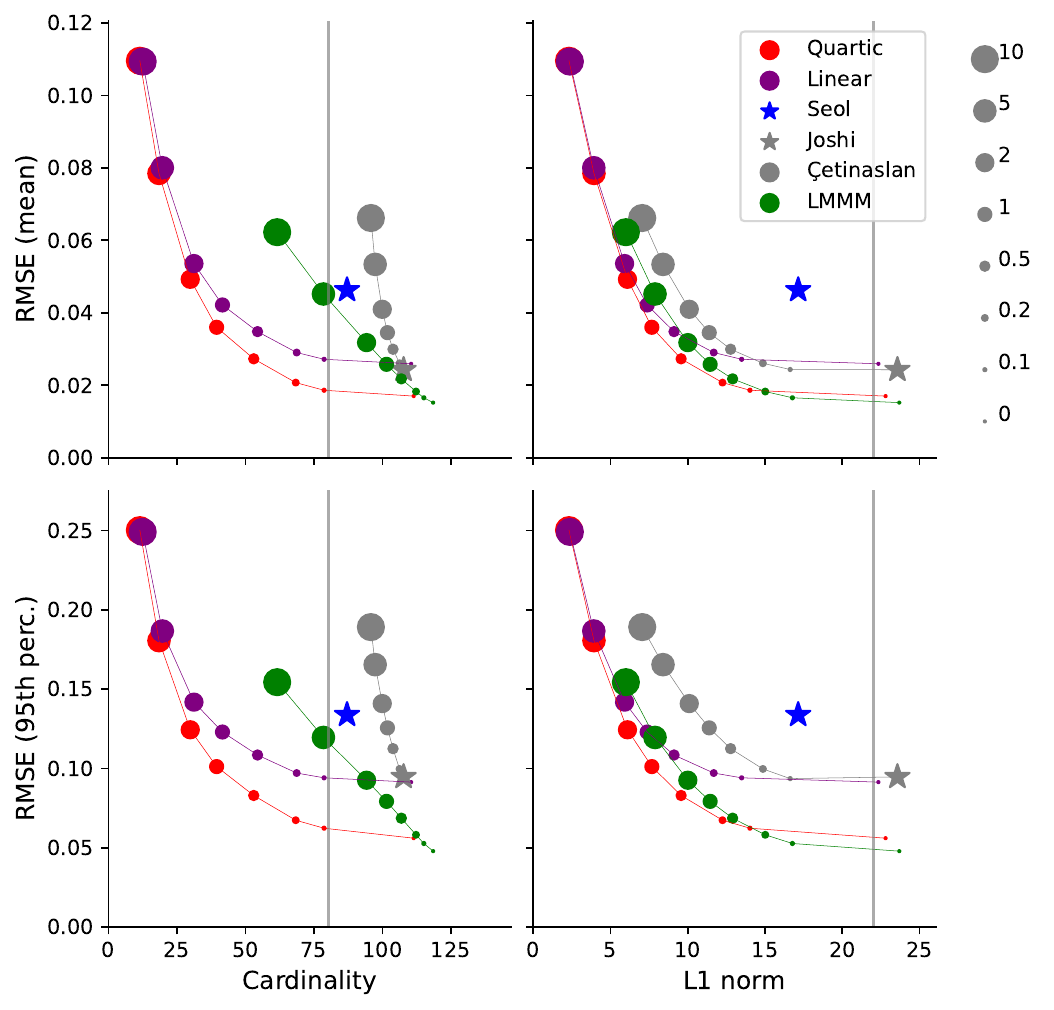}
    \caption{Trade-off between the mesh error and cardinality / L1 norm of the weight vector for \textit{Char 5}. The top row shows the mean mesh error, while the bottom corresponds to the $95^{th}$ percentile of the error. Dot sizes are proportional to the size of regularization parameter $\alpha$ ranging from $0$ to $10$, as indicated on the right.}
    \label{fig:hmTraining}
\end{figure}

In Figure \ref{fig:adaTraining}, we show results for different methods over the training data for \textit{Ada}. Except for the \textit{Seol} and \textit{Joshi}, all the methods include a regularization parameter, hence we need to see how the results behave with varying values of $\alpha$, as indicated by the corresponding sizes of the dots. The top row of the figure shows the mean mesh error, while the bottom gives the $95^{th}$ percentile of the error. An important aspect to notice is that, even though the proposed method (\textit{Quartic} and \textit{Linear}) does not reach as low mesh error as it is possible with \textit{LMMM} or \textit{Çetinaslan}, it does have a favorable shape of the curve --- it offers a nice trade-off between accuracy and cardinality. Imposing higher $\alpha$ values in \textit{Çetinaslan} and \textit{LMMM} increases the mesh error but does not reduce cardinality as much. We can notice a similar behavior for all the other animated characters, in Figures \ref{fig:jesseTraining}--\ref{fig:hmTraining}.

The general conclusions are as follows. \textit{Seol} produces reasonably low cardinality of \textbf{w}, which is around the same value as in artist-crafted reference animation, but relatively high error. \textit{LMMM} can achieve the lowest mesh error of the fit, yet both \textit{LMMM} and \textit{Cetinaslan} have higher cardinality, which cannot be significantly reduced even with relatively high regularization values. The proposed method is the only one that has a good trade-off between mesh error and sparsity of the results, which is tuned by choosing the right $\alpha$ value. We further pick the optimal values of $\alpha$ for each method and proceed to analyze the results in more detail over the test data. Chosen values for each character/method combination are given in Table \ref{tab:AlphaTable}.

\begin{table}[]
    \centering
    \begin{tabular}{c | c  c  c  c  c }
                        & \textit{Ada} & \textit{Jesse} & \textit{Vivian} & \textit{Omar} & \textit{Char 5} \\
                \hline
    \textit{Quartic}    & 0.5          & 1              & 0.5             & 0.5           & 0.5           \\
    \textit{Linear}     & 0.5          & 1              & 0.5             & 0.5           & 0.5           \\        
    \textit{Çetinaslan} & 0.2          & 0.5            & 0.5             & 0.5           & 0.5           \\
       \textit{LMMM}    & 0.2          & 0.5            & 0.5             & 0.5           & 0.5       
    \end{tabular}
    \caption{The selected values of the regularization parameter $\alpha$ for each method and each  character.}
    \label{tab:AlphaTable}
\end{table}

\begin{figure}
    \centering
    \includegraphics[width=0.5\linewidth]{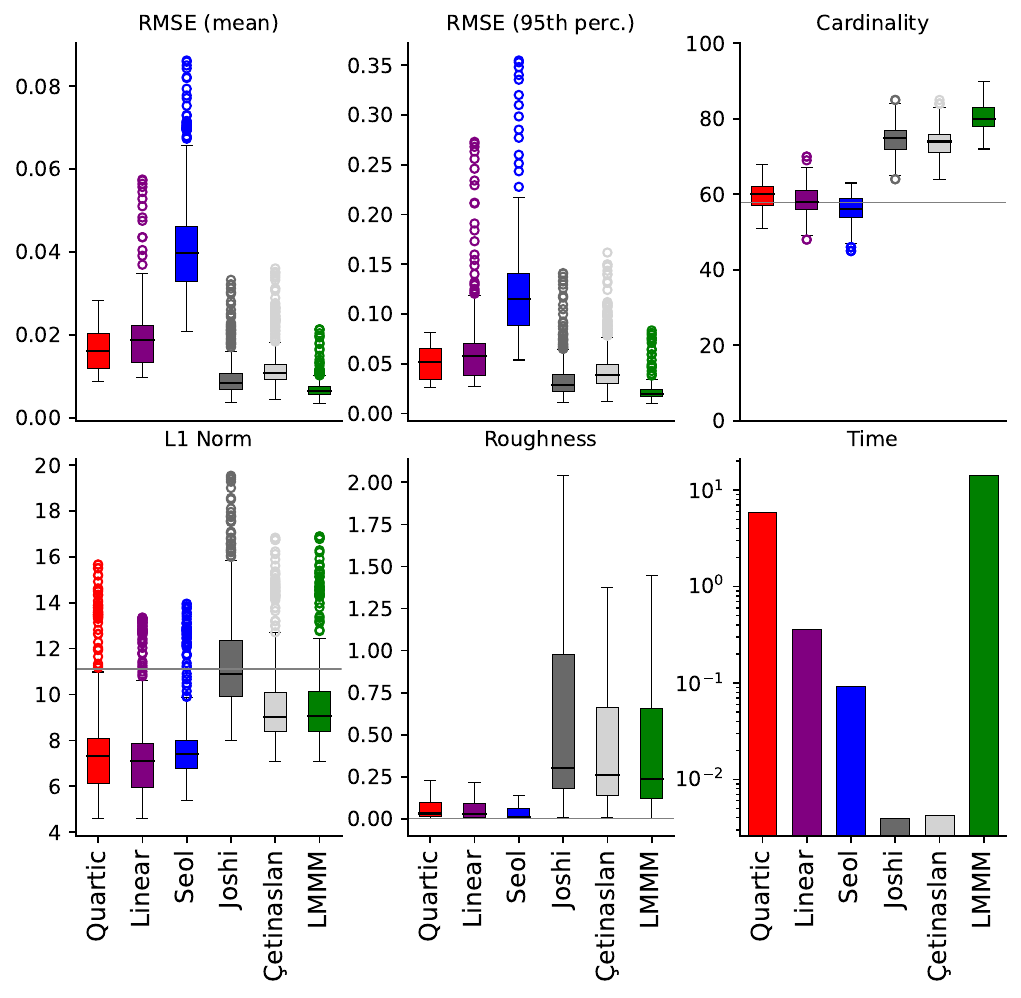}
    \caption{Results statistics for \textit{Ada}, over the test animation. Horizontal gray lines indicate the average value of the corresponding metric in the ground truth data. Execution times are presented in the log scale. For the exact numerical values, consult Table \ref{tab:tabAda}.}
    \label{fig:Adabarplots}
\end{figure}

Further, we proceed with the corresponding values and evaluate all the methods over the test data. For the test case, we consider animation sequences (see supplementary video materials), which allows us also to estimate the temporal smoothness of the results, as explained in Section \ref{sed:metrics}. The resulting metric values for \textit{Ada} are presented in Figure \ref{fig:Adabarplots}, and accompanied by Table \ref{tab:tabAda}. Figure \ref{fig:Adabarplots} shows a visible separation between the coordinate descent-based methods (\textit{Quartic}, \textit{Linear} and \textit{Seol}) that lead to low cardinality and smooth results, versus methods that estimate the weights jointly (\textit{Joshi}, \textit{Çetinaslan} and \textit{LMMM}), which lead to low RMSE but a more dense vector of weights. Notice that our method \textit{Quartic} gives relatively low RMSE, comparable to those of \textit{LMMM} and \textit{Çetinaslan}, when observing both the mean error and the $95^{th}$ percentile. At the same time, it gives sparse results, comparable with a less accurate method of \textit{Seol}, and smooth frame-to-frame transitions. The only weak point of our method is the execution time, yet it is still only a third of the time needed for the \textit{LMMM} solution. In other words, we can cope with more complex models like \textit{LMMM} but with an approximately three times faster solution.

Figure 1 gives an example frame for \textit{Ada}, comparing the six methods. The top row shows reconstructed meshes, where red tones indicate the regions of higher error. The bottom row shows the corresponding blendshape activation weights (sorted by the weights of the reference frame). One can notice that \textit{Seol} gives a higher reconstruction error, producing a completely wrong facial expression. \textit{Linear}, \textit{Joshi} and \textit{Çetinaslan} are better, yet the lower lip is slightly off.  \textit{Quartic} and \textit{LMMM} give very accurate mesh reconstruction, but our method produces a considerably sparser vector of weights than \textit{LMMM}.

More visual examples are given in Figure \ref{fig:details_ada}. For visualization purposes, we excluded \textit{Linear} and \textit{Joshi}, as less significant, and we zoomed in on the mouth regions since that is where most of the visible error is. All the examples confirm the previously stated conclusions that \textit{Quartic} and \textit{LMMM} lead to an accurate mesh reconstruction, while only our method gives an optimal trade-off between the mesh error and the sparsity of the weight vector. At the same time, the execution time of the proposed method is three times lower than that of \textit{LMMM} (the only benchmark without visible mesh artifacts), and the value of \textit{Roughness} metric is drastically lower compared to any of the benchmarks that solve the weights jointly.

\begin{table}[]
    \centering
    \begin{tabular}{c | c c c c c c}
            &\makecell{RMSE \\ mean }&\makecell{RMSE \\ $95^{th}$} & Card.&\makecell{L1 \\ norm}& Rough.         & Time              \\
                \hline
        Quartic      & 0.015        & 0.050        & 59.3              & 7.68              & 0.108              & 5.848               \\
        Linear       & 0.019        & 0.061        & 58.1              &\happy\textbf{7.37}& 0.108              & 0.361               \\        
        Seol         & \done 0.041  & \done 0.120  &\happy\textbf{55.9}& 7.79              &\happy\textbf{0.092}& 0.091               \\
        Joshi        & 0.009        & 0.034        & 74.7              & \done 11.3        & \done 2.838        &\happy\textbf{0.004} \\
        Çetinaslan   & 0.012        & 0.044        & 74.0              & 9.56              & 0.583              & 0.004               \\
        LMMM         &\happy\textbf{0.007}&\happy\textbf{0.021}& \done 80.2& 9.57          & 0.567              & \done 14.09
    \end{tabular}
    \caption{\textit{Ada}. Average values for each metric and each method, corresponding to Figure \ref{fig:Adabarplots}. The worst score for each column is shaded, while the best is highlighted and bold.}
    \label{tab:tabAda}
\end{table}

In the supplemental video materials, a reader can better grasp the visual differences since the animated sequence is represented side-by-side with a reconstruction of each method. The proposed method with quartic rig function, and \textit{LMMM}, give almost flawless reconstructions, while \textit{Seol} is visibly worse than any of the other methods. An interesting aspect to notice in the case of \textit{Joshi} and \textit{Çetinaslan} is that they produce a shivering-like effect throughout the entire sequence (especially visible in the lips). This aligns with the exhibited high value for roughness in Figure \ref{fig:Adabarplots}, i.e., even though individual frames give a relatively good mesh reconstruction, weight activations will differ more significantly between the consecutive frames. This artifact can also be noticed in \textit{LMMM}, although it is very subtle and easy to oversee. On the other side, our method gives both a good reconstruction in the individual frames and temporally smooth animation. 

\begin{figure}
    \centering
    \includegraphics[width=0.5\linewidth]{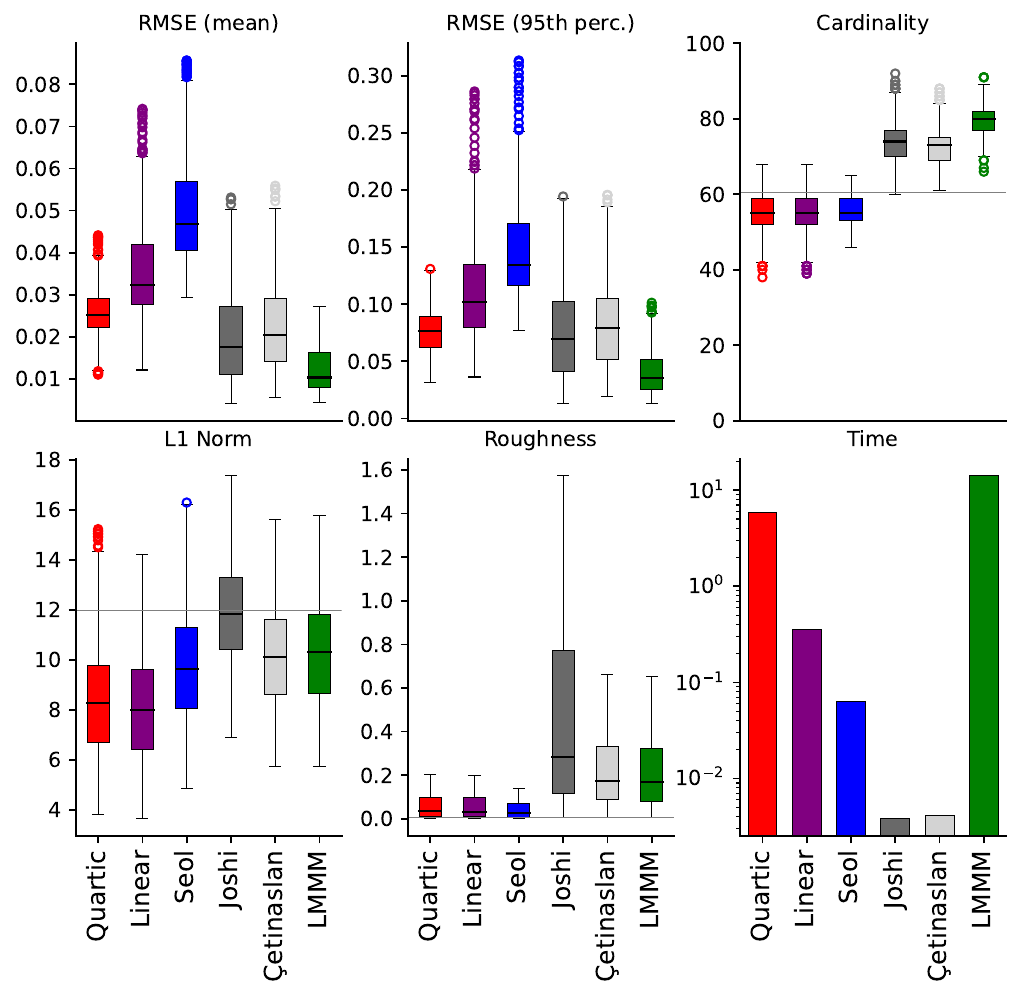}
    \caption{Results statistics for \textit{Jesse}, over the test animation. Horizontal gray lines indicate the average value of the corresponding metric in the ground truth data. Execution times are presented in the log scale. For the exact numerical values, consult Table \ref{tab:tabJesse}.}
    \label{fig:Jessebarplots}
\end{figure}

\begin{figure}
    \centering
    \includegraphics[width=0.5\linewidth]{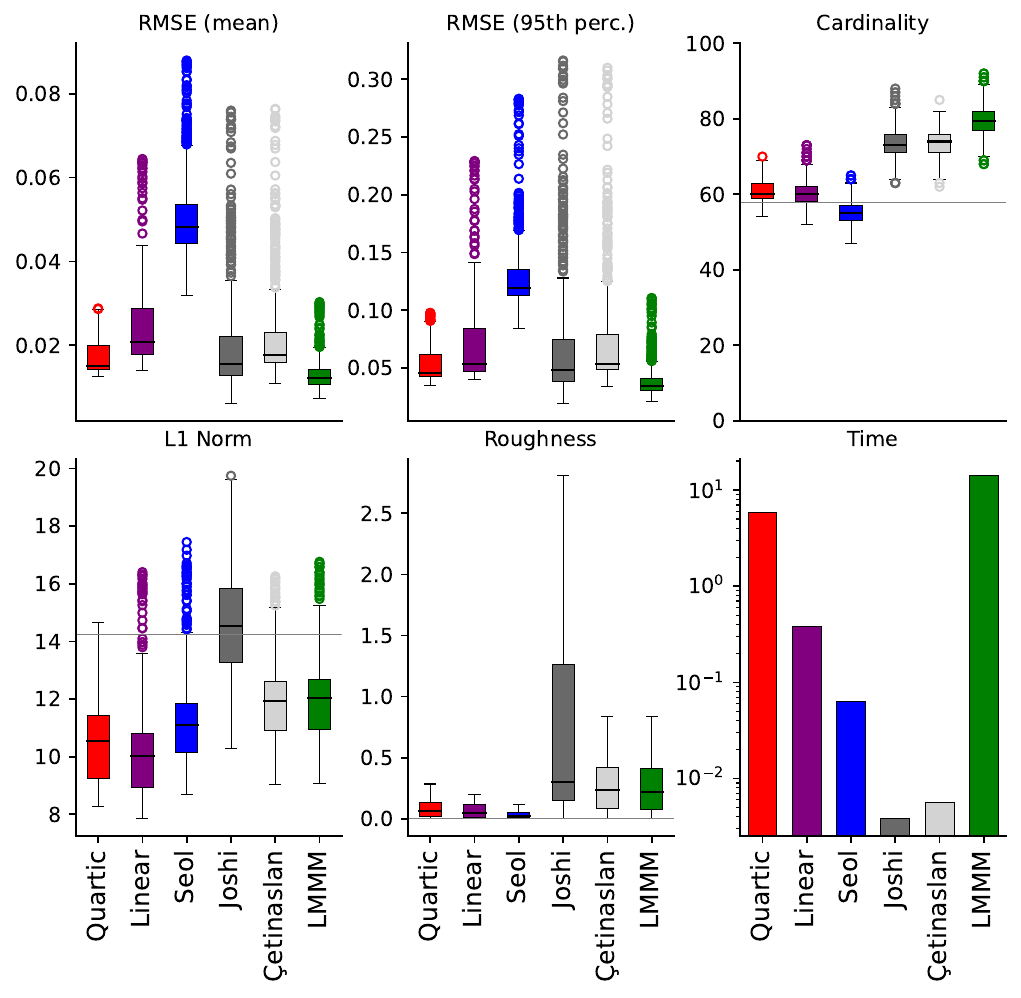}
    \caption{Results statistics for \textit{Vivian}, over the test animation. Horizontal gray lines indicate the average value of the corresponding metric in the ground truth data. Execution times are presented in the log scale. For the exact numerical values, consult Table \ref{tab:tabVivian}.}
    \label{fig:Vivianbarplots}
\end{figure}

\begin{figure}
    \centering
    \includegraphics[width=0.5\linewidth]{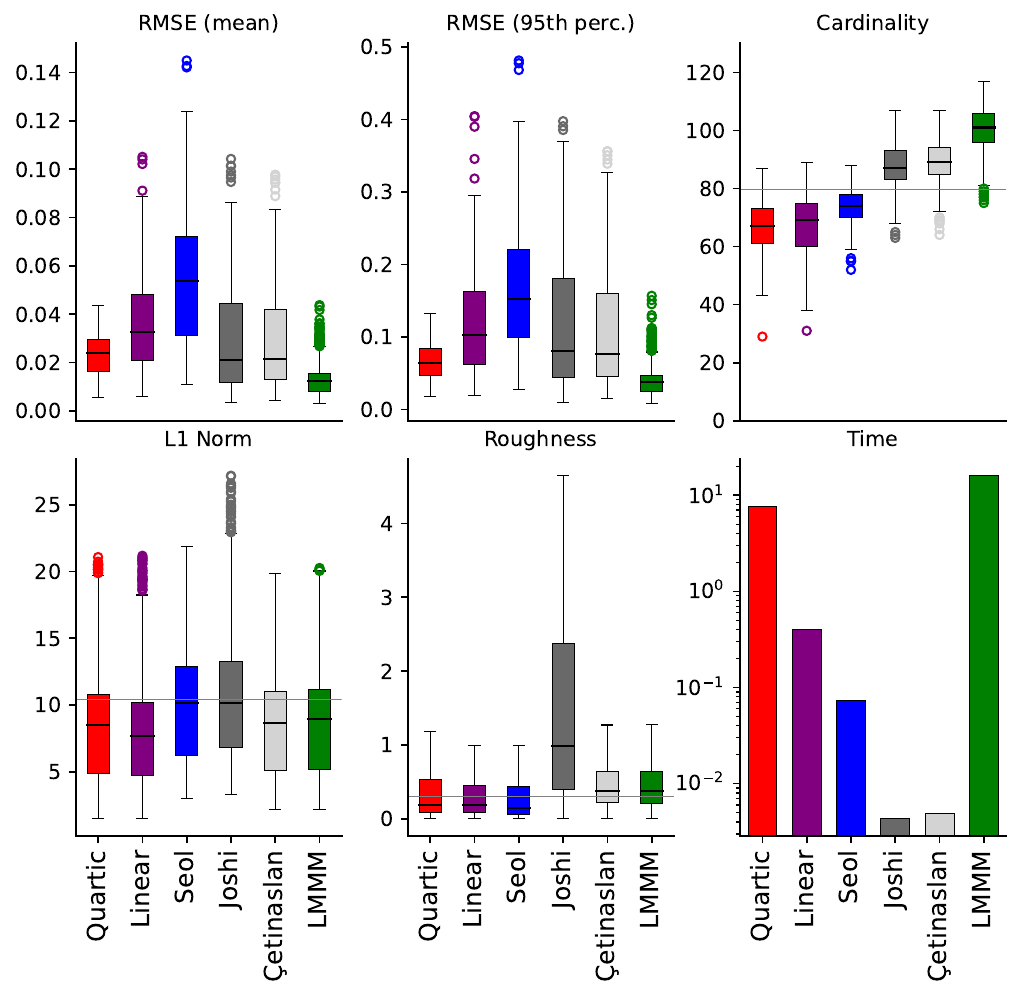}
    \caption{Results statistics for \textit{Omar}, over the test animation. Horizontal gray lines indicate the average value of the corresponding metric in the ground truth data. Execution times are presented in the log scale. For the exact numerical values, consult Table \ref{tab:tabOmar}.}
    \label{fig:Omarbarplots}
\end{figure}

\begin{figure}
    \centering
    \includegraphics[width=0.5\linewidth]{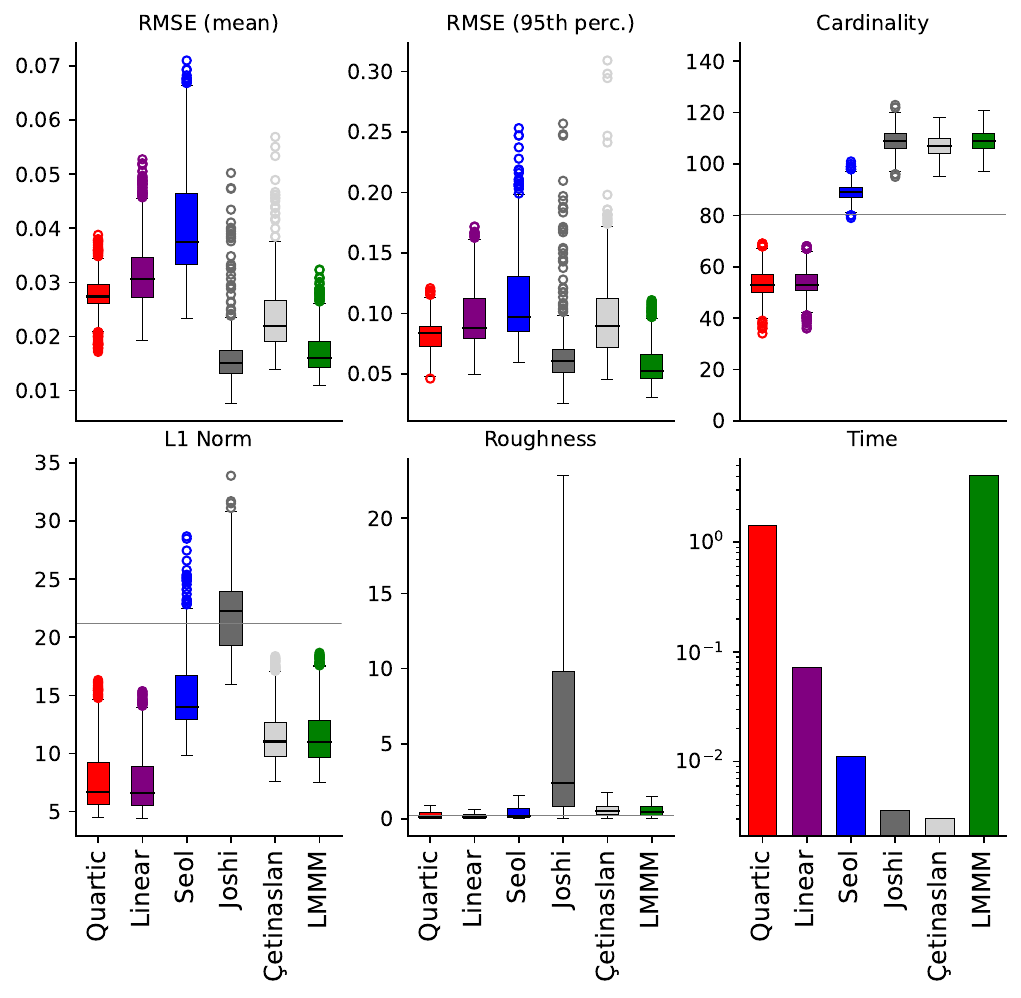}
    \caption{Results statistics for \textit{Char 5}, over the test animation. Horizontal gray lines indicate the average value of the corresponding metric in the ground truth data. Execution times are presented in the log scale. For the exact numerical values, consult Table \ref{tab:tabHM}.}
    \label{fig:HMbarplots}
\end{figure}

\begin{figure}
\centering
  \setlength{\unitlength}{0.1\textwidth}
          \begin{tikzpicture}
        \node[above right, inner sep=0] (image) at (  0,  7.7){\includegraphics[width=0.5\linewidth]{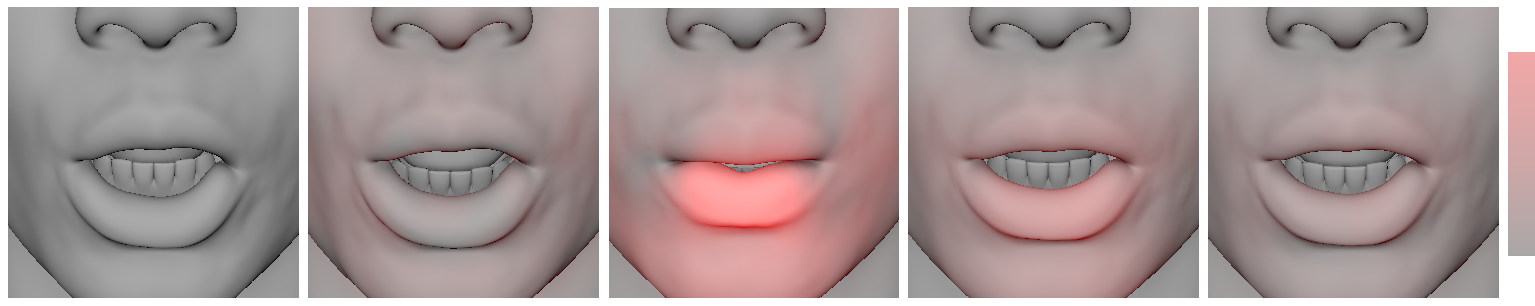}};
        \node[above right, inner sep=0] (image) at (-0.05,6.2){\includegraphics[width=0.5\linewidth]{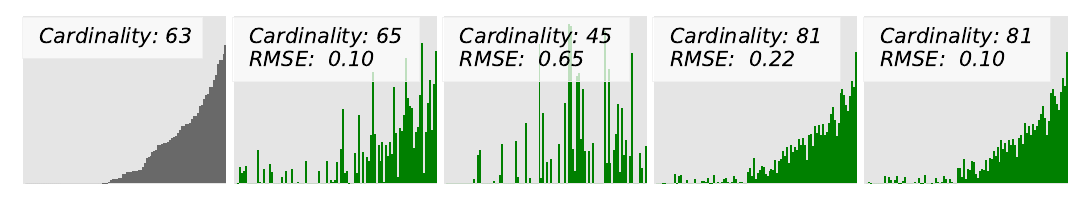}};
        \node[above right, inner sep=0] (image) at (  0,  4.7){\includegraphics[width=0.5\linewidth]{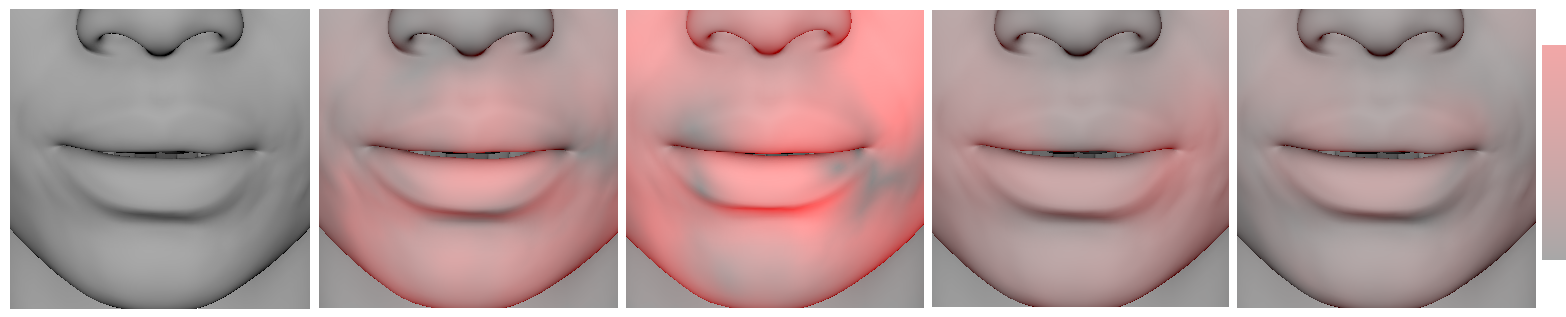}};
        \node[above right, inner sep=0] (image) at (-0.05,3.2){\includegraphics[width=0.5\linewidth]{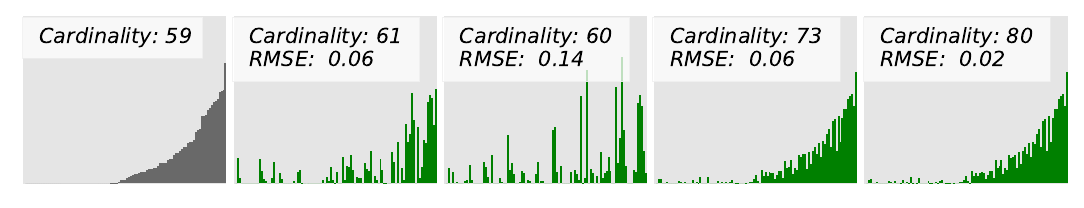}};   \node[above right, inner sep=0] (image) at (  0,  1.7){\includegraphics[width=0.5\linewidth]{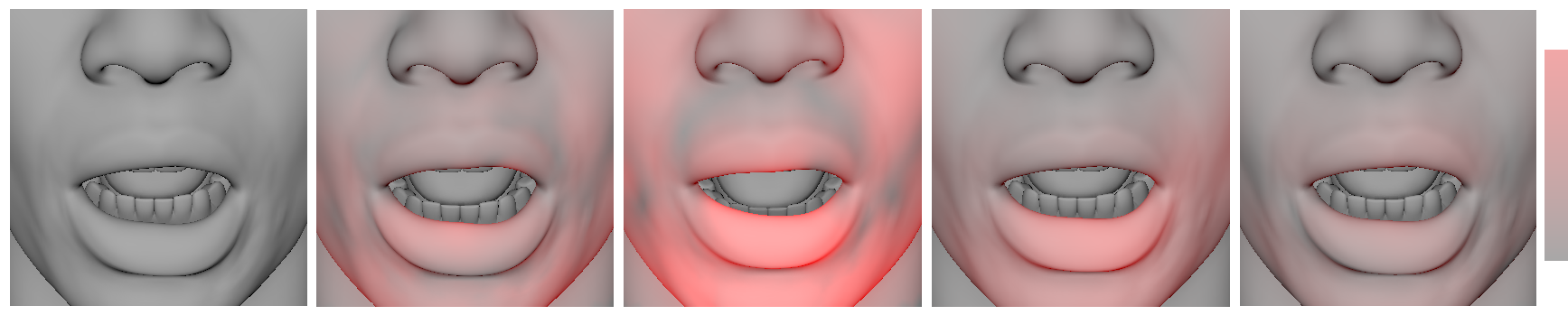}};
        \node[above right, inner sep=0] (image) at (-0.05,0.2){\includegraphics[width=0.5\linewidth]{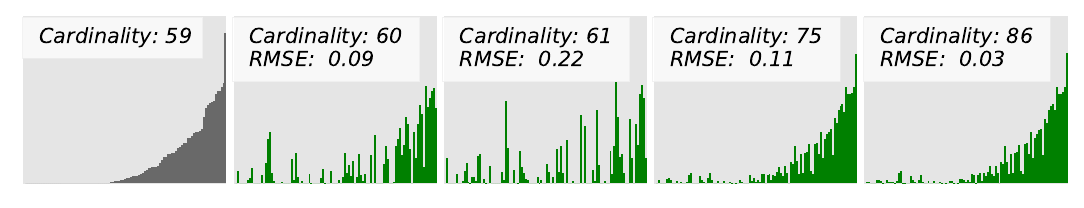}};   \begin{scope}[
        x={($0.1*(image.south east)$)},
        y={($0.1*(image.north west)$)}]
            \node[darkgray] at  (0.9,1){\small Reference };
            \node[darkgray] at  (3.0,0.75){\small Quartic (ours) };
            \node[darkgray] at  (4.95,0.5){\small Seol };
            \node[darkgray] at  (7.0,0.25){\small Çetinaslan};
            \node[darkgray] at  (9.0,0.){\small LMMM };
            \node[darkgray] at  (10.30,50){\scriptsize .66 };
            \node[darkgray] at  (10.25,42.2){\scriptsize .00 };
            \node[darkgray] at  (10.23,33.5){\scriptsize .15 };
            \node[darkgray] at  (10.18,26){\scriptsize .00 };
            \node[darkgray] at  (10.10,17.1){\scriptsize .22 };
            \node[darkgray] at  (10.05,9){\scriptsize .00 };
        \end{scope}
        \end{tikzpicture}
  \caption{\textit{Ada}, example frames with predictions using different methods. The odd rows show mesh reconstructions, and regions of higher mesh error are highlighted in red, according to the color bar on the right. The even rows show corresponding blendshape weights activations, with summarized root mean squared error and cardinality of each approach.}
    \label{fig:details_ada}
\end{figure}

\begin{figure*}
  \centering
  \setlength{\unitlength}{0.1\textwidth}
          \begin{tikzpicture}
        \node[above right, inner sep=0] (image) at (0,2.55){\includegraphics[width=0.975\textwidth]{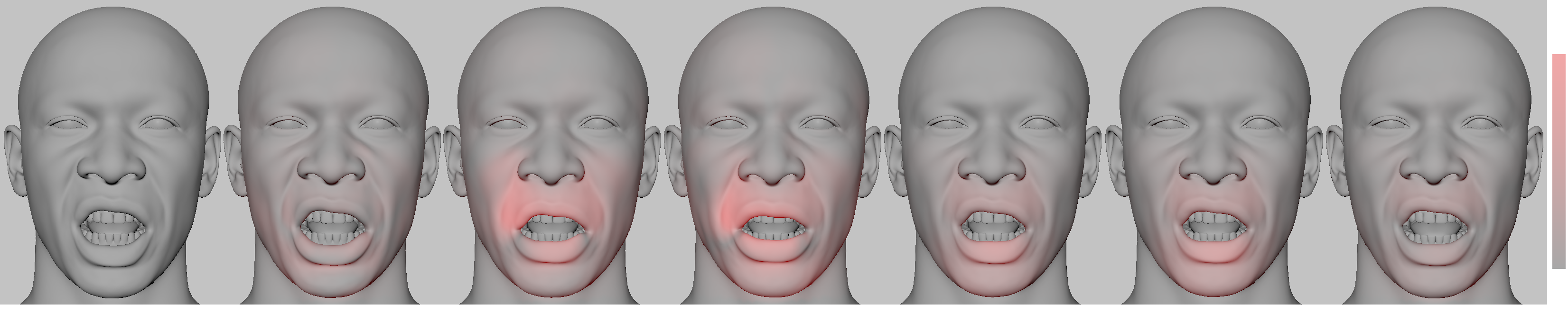}};
        \node[above right, inner sep=0] (image) at (-0.2,0){\includegraphics[width=0.975\textwidth]{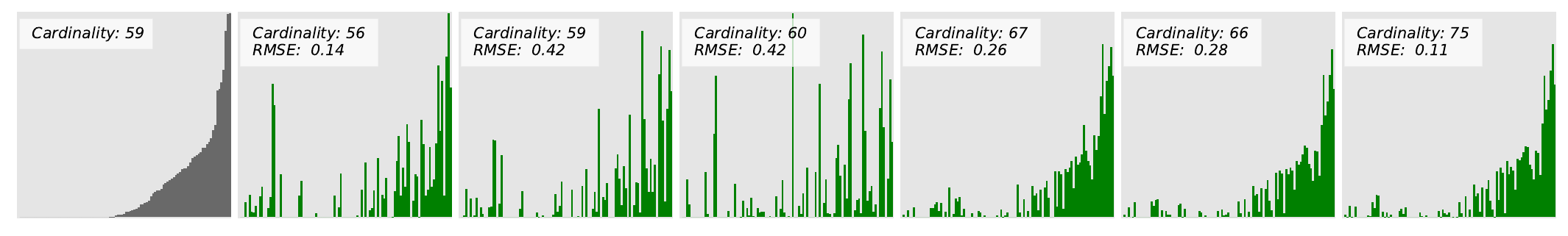}};
        \begin{scope}[
        x={($0.1*(image.south east)$)},
        y={($0.1*(image.north west)$)}]
            \node[darkgray] at  (0.70,0.2){\small Reference };
            \node[darkgray] at  (2.10,0.2){\small Quartic (ours) };
            \node[darkgray] at  (3.50,0.2){\small Linear (ours) };
            \node[darkgray] at  (4.95,0.2){\small Seol };
            \node[darkgray] at  (6.30,0.2){\small Joshi };
            \node[darkgray] at  (7.80,0.2){\small Çetinaslan };
            \node[darkgray] at  (9.20,0.2){\small LMMM };
            \node[darkgray] at  (10.375,22.5){\scriptsize .43 };
            \node[darkgray] at  (10.24,12){\scriptsize .00 };
            \node[darkgray] at  (10.25,11){\scriptsize cm };
        \end{scope}
        \end{tikzpicture}
  \caption{\textit{Jesse}, an example frame with predictions using different methods. The top row shows a mesh reconstruction, and regions of higher mesh error are highlighted in red, according to the color bar on the right. The bottom row shows corresponding blendshape weights activation, with summarized root mean squared error and cardinality of each approach.}
  \label{fig:wide2}
\end{figure*}

\begin{figure}
\centering
  \setlength{\unitlength}{0.1\textwidth}
          \begin{tikzpicture}
        \node[above right, inner sep=0] (image) at (  0,  7.7){\includegraphics[width=0.5\linewidth]{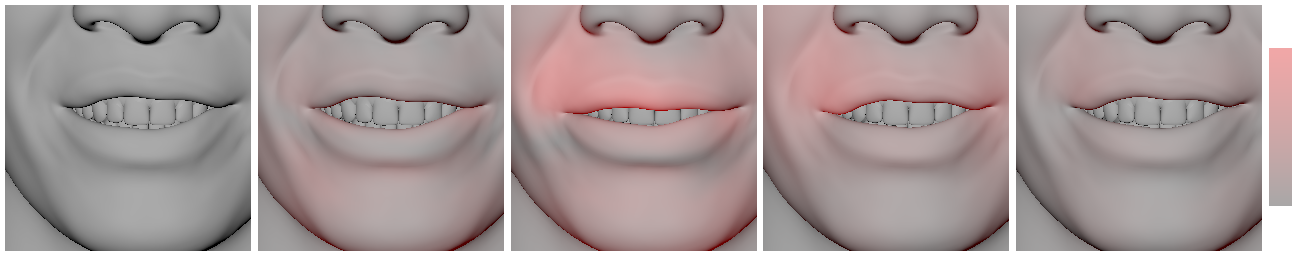}};
        \node[above right, inner sep=0] (image) at (-0.05,6.2){\includegraphics[width=0.5\linewidth]{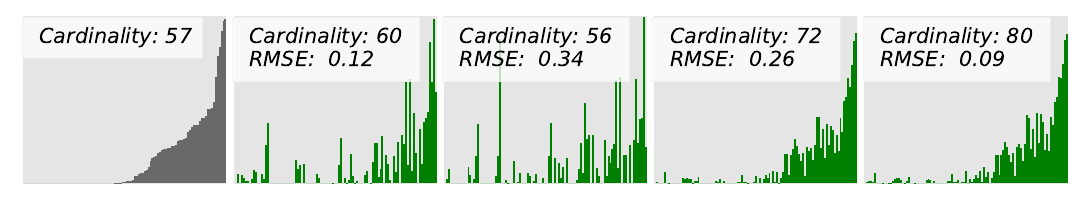}};
        \node[above right, inner sep=0] (image) at (  0,  4.7){\includegraphics[width=0.5\linewidth]{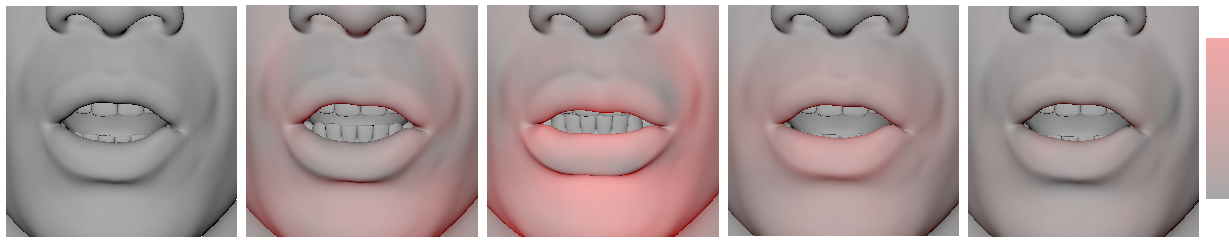}};
        \node[above right, inner sep=0] (image) at (-0.05,3.2){\includegraphics[width=0.5\linewidth]{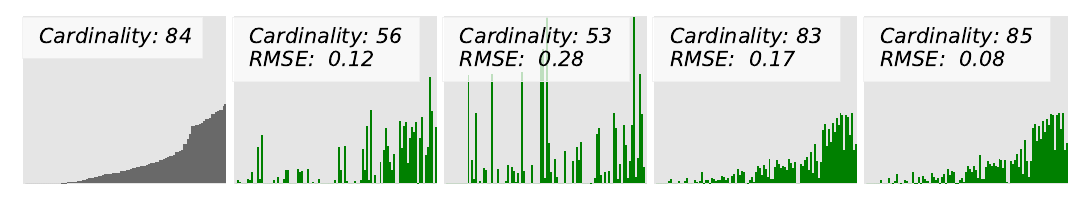}};
        \node[above right, inner sep=0] (image) at (  0,  1.7){\includegraphics[width=0.5\linewidth]{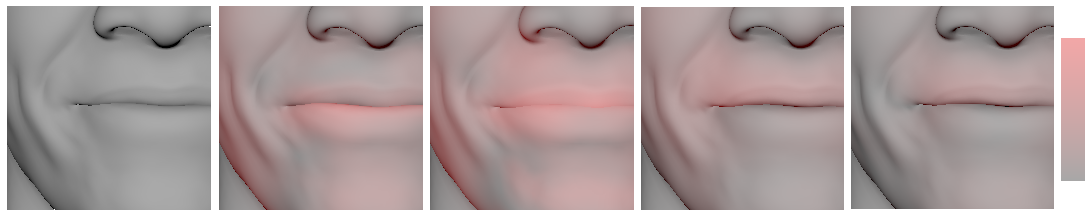}};
        \node[above right, inner sep=0] (image) at (-0.05,0.2){\includegraphics[width=0.5\linewidth]{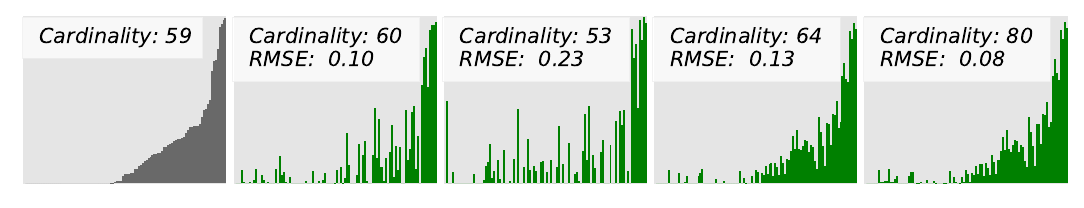}};   \begin{scope}[
        x={($0.1*(image.south east)$)},
        y={($0.1*(image.north west)$)}]
            \node[darkgray] at  (0.9,1.){\small Reference };
            \node[darkgray] at  (3.0,0.75){\small Quartic (ours) };
            \node[darkgray] at  (4.95,0.5){\small Seol };
            \node[darkgray] at  (7.0,0.25){\small Çetinaslan};
            \node[darkgray] at  (9.0,0.){\small LMMM };
            \node[darkgray] at  (10.3, 50){\scriptsize .39 };
            \node[darkgray] at  (10.25,42.5){\scriptsize .00 };
            \node[darkgray] at  (10.2, 33.5){\scriptsize .28 };
            \node[darkgray] at  (10.15,26){\scriptsize .00 };
            \node[darkgray] at  (10.1, 16.9){\scriptsize .24 };
            \node[darkgray] at  (10.05,9){\scriptsize .00 };
        \end{scope}
        \end{tikzpicture}
  \caption{\textit{Jesse}, example frames with predictions using different methods. The odd rows show mesh reconstructions, and regions of higher mesh error are highlighted in red, according to the color bar on the right. The even rows show corresponding blendshape weights activations, with summarized root mean squared error and cardinality of each approach.}
    \label{fig:details_jesse}
\end{figure}

\begin{figure*}
  \centering
  \setlength{\unitlength}{0.1\textwidth}
          \begin{tikzpicture}
        \node[above right, inner sep=0] (image) at (0,2.6){\includegraphics[width=0.975\textwidth]{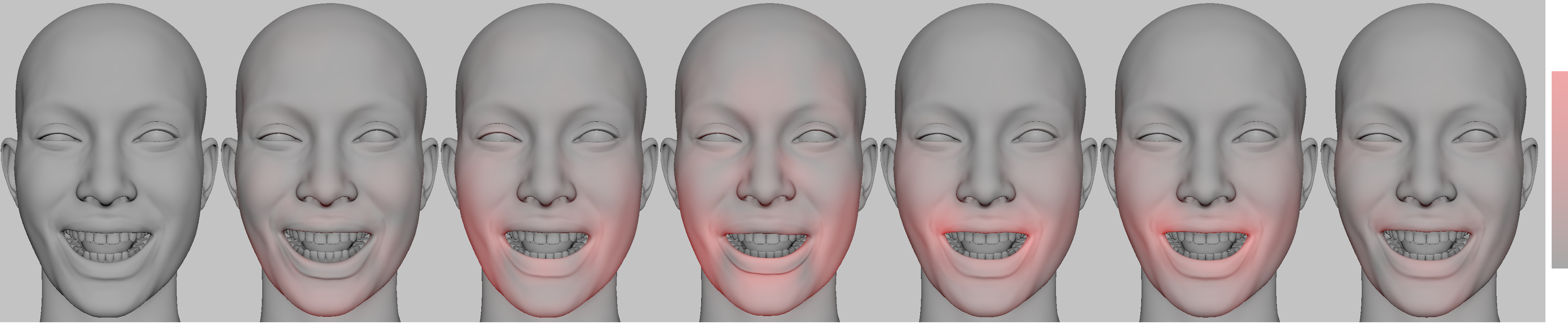}};
        \node[above right, inner sep=0] (image) at (-0.2,0){\includegraphics[width=0.975\textwidth]{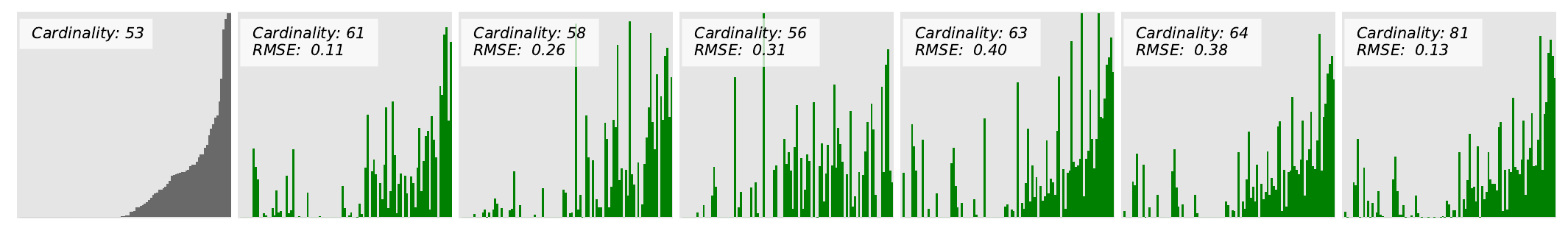}};
        \begin{scope}[
        x={($0.1*(image.south east)$)},
        y={($0.1*(image.north west)$)}]
            \node[darkgray] at  (0.70,0.2){\small Reference };
            \node[darkgray] at  (2.10,0.2){\small Quartic (ours) };
            \node[darkgray] at  (3.50,0.2){\small Linear (ours) };
            \node[darkgray] at  (4.95,0.2){\small Seol };
            \node[darkgray] at  (6.30,0.2){\small Joshi };
            \node[darkgray] at  (7.80,0.2){\small Çetinaslan };
            \node[darkgray] at  (9.20,0.2){\small LMMM };
            \node[darkgray] at  (10.35,22){\footnotesize .40 };
            \node[darkgray] at  (10.25,12){\footnotesize .00 };
            \node[darkgray] at  (10.25,11){\footnotesize cm };
        \end{scope}
        \end{tikzpicture}
  \caption{\textit{Vivian}, an example frame with predictions using different methods. The top row shows a mesh reconstruction, and regions of higher mesh error are highlighted in red, according to the color bar on the right. The bottom row shows corresponding blendshape weights activation, with summarized root mean squared error and cardinality of each approach.}
  \label{fig:wide3}
\end{figure*}

\begin{figure}
\centering
  \setlength{\unitlength}{0.1\textwidth}
          \begin{tikzpicture}
        \node[above right, inner sep=0] (image) at (   0, 7.7){\includegraphics[width=0.5\linewidth]{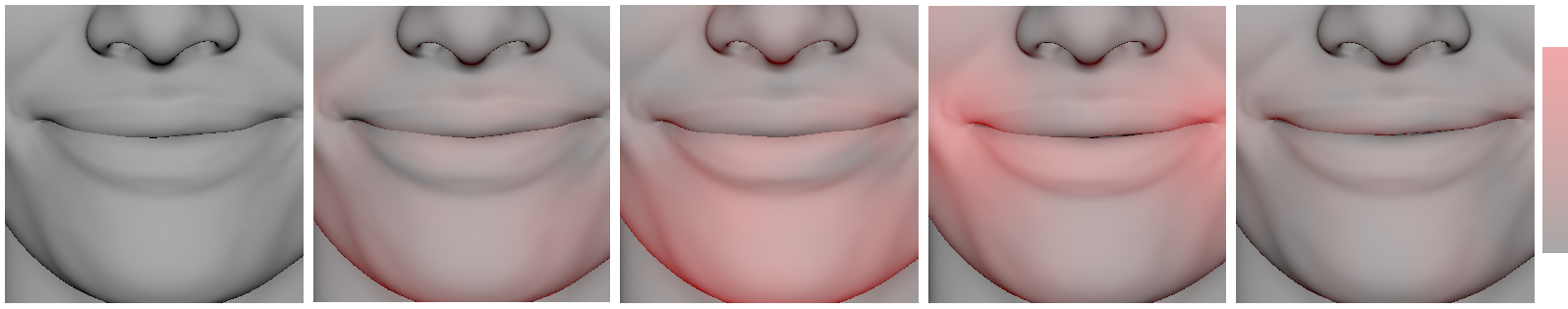}};
        \node[above right, inner sep=0] (image) at (-0.05,6.2){\includegraphics[width=0.5\linewidth]{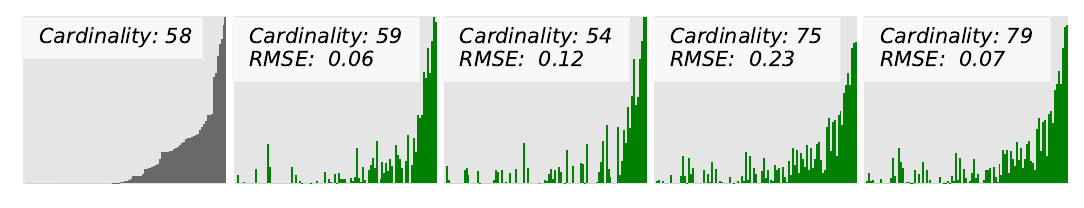}};
        \node[above right, inner sep=0] (image) at (   0, 4.7){\includegraphics[width=0.5\linewidth]{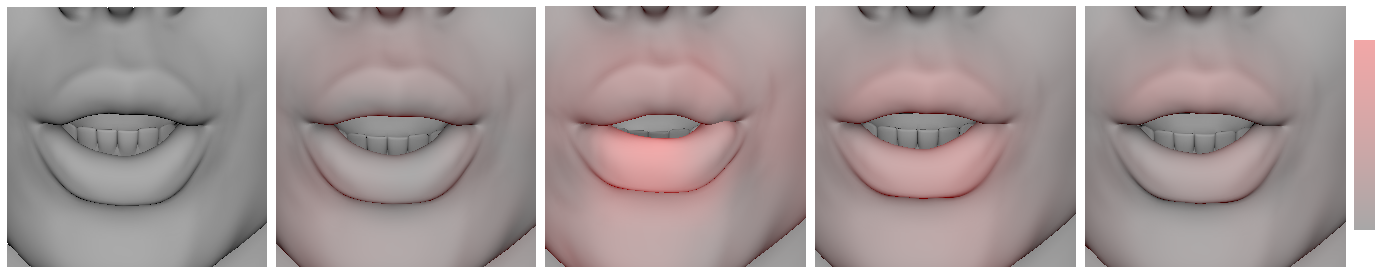}};
        \node[above right, inner sep=0] (image) at (-0.05,3.2){\includegraphics[width=0.5\linewidth]{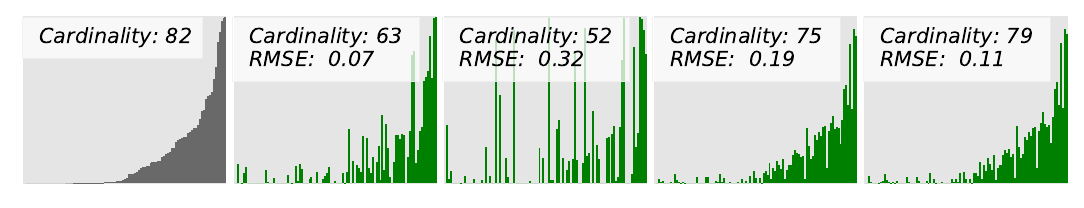}};  \node[above right, inner sep=0] (image) at (   0, 1.7){\includegraphics[width=0.5\linewidth]{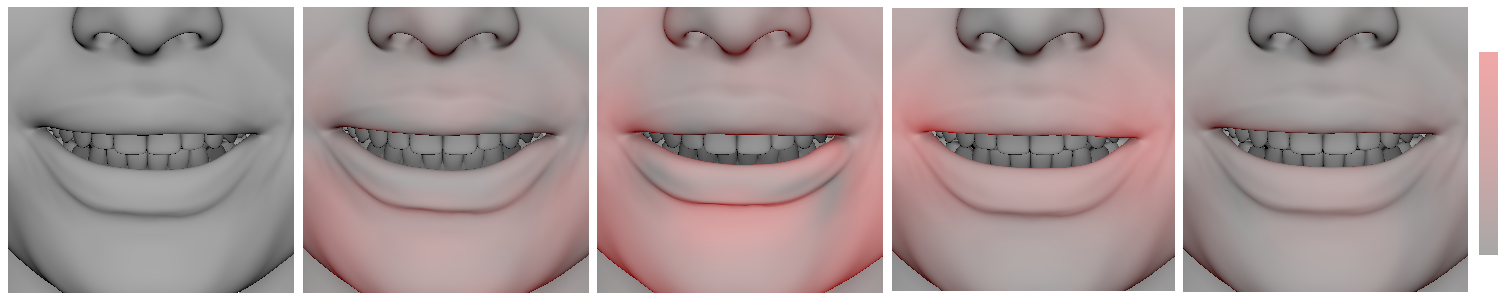}};
        \node[above right, inner sep=0] (image) at (-0.05,0.2){\includegraphics[width=0.5\linewidth]{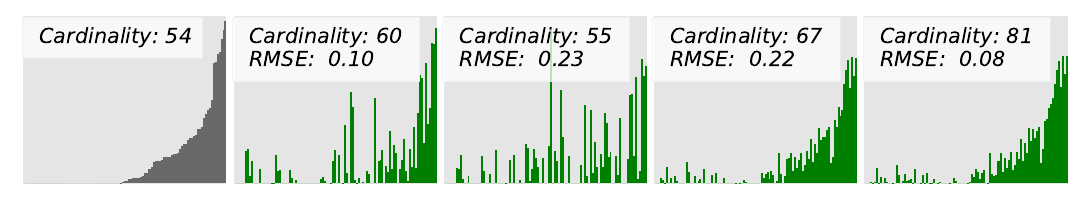}};   \begin{scope}[
        x={($0.1*(image.south east)$)},
        y={($0.1*(image.north west)$)}]
            \node[darkgray] at  (0.9,1.){\small Reference };
            \node[darkgray] at  (3.0,0.75){\small Quartic (ours) };
            \node[darkgray] at  (4.95,0.5){\small Seol };
            \node[darkgray] at  (7.0,0.25){\small Çetinaslan};
            \node[darkgray] at  (9.0,0.){\small LMMM };
            \node[darkgray] at  (10.3, 49.9){\scriptsize .23 };
            \node[darkgray] at  (10.25,42.5){\scriptsize .00 };
            \node[darkgray] at  (10.2, 33.5){\scriptsize .32 };
            \node[darkgray] at  (10.15,25.7){\scriptsize .00 };
            \node[darkgray] at  (10.1, 16.9){\scriptsize .26 };
            \node[darkgray] at  (10.05,9){\scriptsize .00 };
        \end{scope}
        \end{tikzpicture}
  \caption{\textit{Vivian}, example frames with predictions using different methods. The odd rows show mesh reconstructions, and regions of higher mesh error are highlighted in red, according to the color bar on the right. The even rows show corresponding blendshape weights activations, with summarized root mean squared error and cardinality of each approach.}
    \label{fig:details_vivian}
\end{figure}

\begin{figure*}
  \centering
  \setlength{\unitlength}{0.1\textwidth}
          \begin{tikzpicture}
        \node[above right, inner sep=0] (image) at (0,2.6){\includegraphics[width=0.975\textwidth]{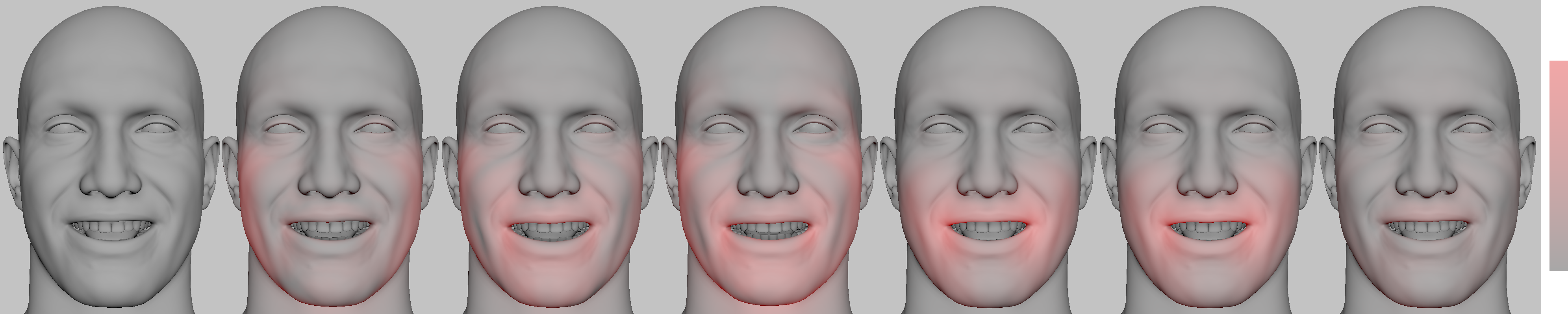}};
        \node[above right, inner sep=0] (image) at (-0.2,0){\includegraphics[width=0.975\textwidth]{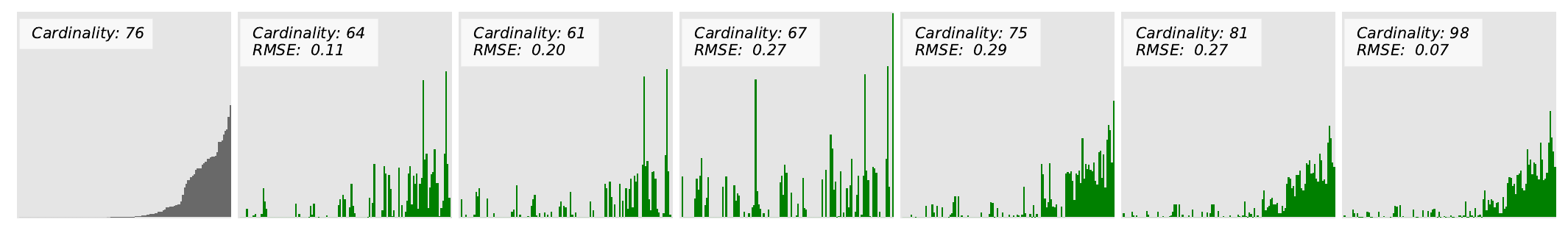}};
        \begin{scope}[
        x={($0.1*(image.south east)$)},
        y={($0.1*(image.north west)$)}]
            \node[darkgray] at  (0.70,0.2){\small Reference };
            \node[darkgray] at  (2.10,0.2){\small Quartic (ours) };
            \node[darkgray] at  (3.50,0.2){\small Linear (ours) };
            \node[darkgray] at  (4.95,0.2){\small Seol };
            \node[darkgray] at  (6.30,0.2){\small Joshi };
            \node[darkgray] at  (7.80,0.2){\small Çetinaslan };
            \node[darkgray] at  (9.20,0.2){\small LMMM };
            \node[darkgray] at  (10.35,22.5){\footnotesize .30 };
            \node[darkgray] at  (10.2,11.5){\footnotesize .00 };
            \node[darkgray] at  (10.2,10.5){\footnotesize cm };
        \end{scope}
        \end{tikzpicture}
  \caption{\textit{Omar}, an example frame with predictions using different methods. The top row shows a mesh reconstruction, and regions of higher mesh error are highlighted in red, according to the color bar on the right. The bottom row shows corresponding blendshape weights activation, with summarized root mean squared error and cardinality of each approach.}
  \label{fig:wide4}
\end{figure*}

\begin{figure}
\centering
  \setlength{\unitlength}{0.1\textwidth}
          \begin{tikzpicture}
        \node[above right, inner sep=0] (image) at (    0,7.7){\includegraphics[width=0.5\linewidth]{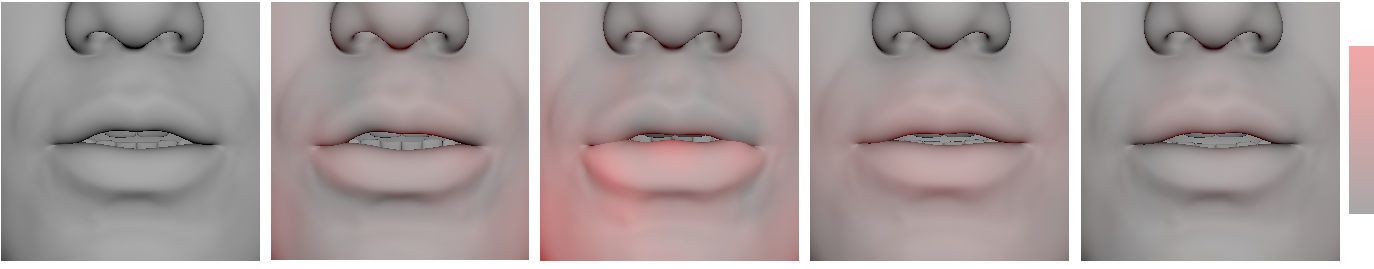}};
        \node[above right, inner sep=0] (image) at (-0.05,6.2){\includegraphics[width=0.5\linewidth]{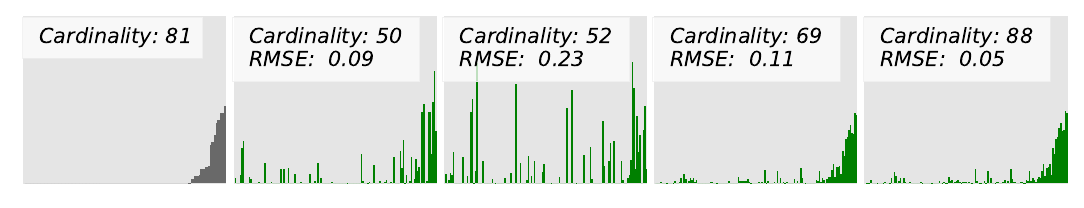}};
        \node[above right, inner sep=0] (image) at (    0,4.7){\includegraphics[width=0.5\linewidth]{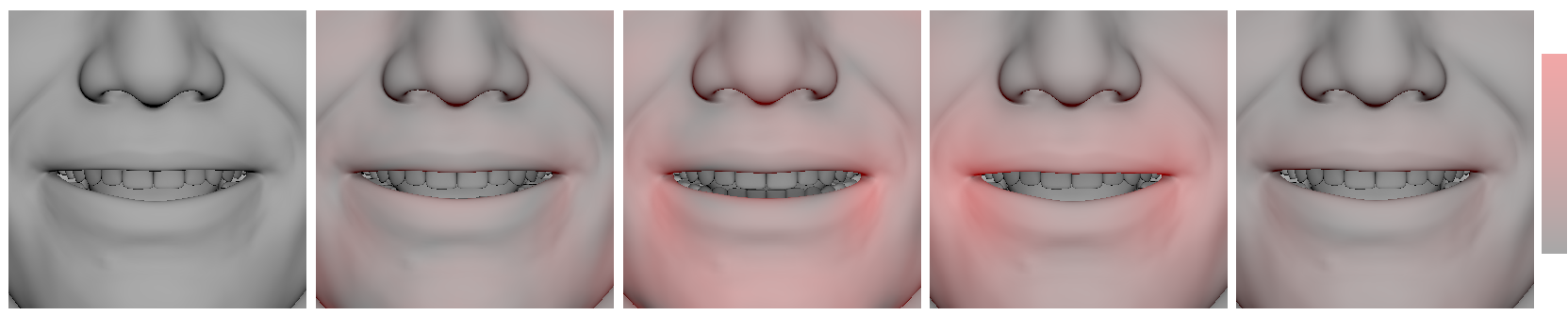}};
        \node[above right, inner sep=0] (image) at (-0.05,3.2){\includegraphics[width=0.5\linewidth]{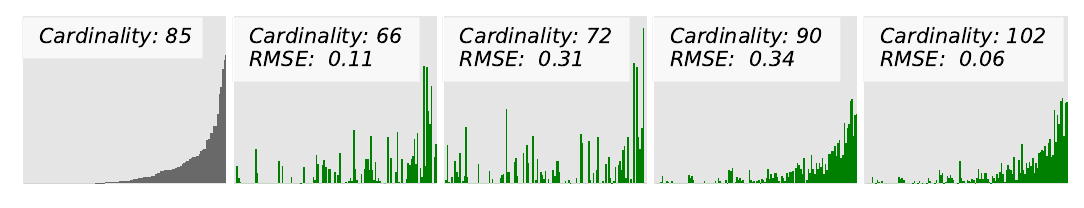}};    \node[above right, inner sep=0] (image) at (    0,1.7){\includegraphics[width=0.5\linewidth]{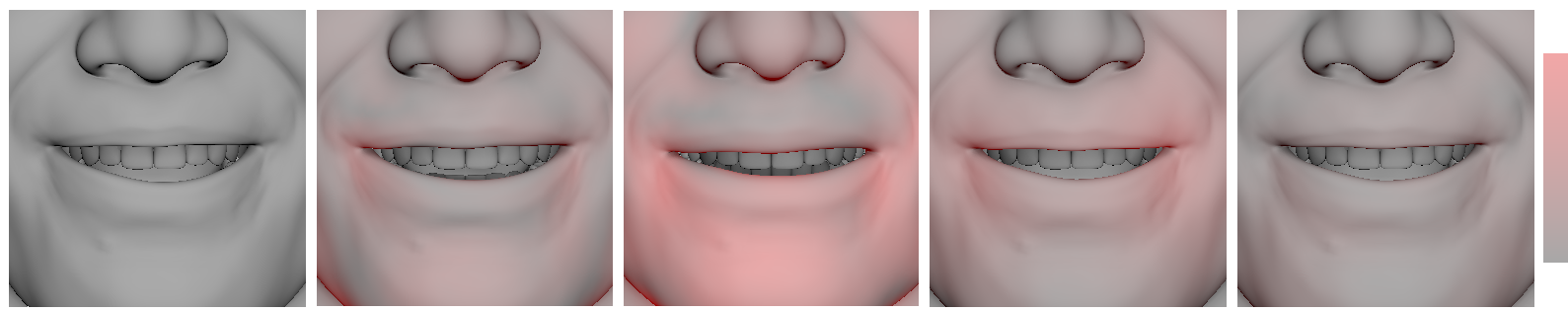}};
        \node[above right, inner sep=0] (image) at (-0.05,0.2){\includegraphics[width=0.5\linewidth]{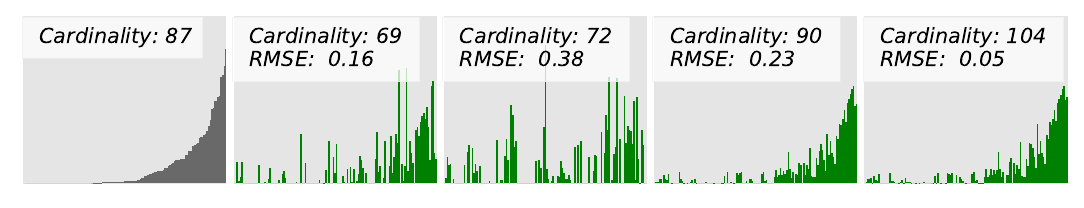}};   \begin{scope}[
        x={($0.1*(image.south east)$)},
        y={($0.1*(image.north west)$)}]
            \node[darkgray] at  (0.9,1.){\small Reference };
            \node[darkgray] at  (3.0,0.75){\small Quartic (ours) };
            \node[darkgray] at  (4.95,0.5){\small Seol };
            \node[darkgray] at  (7.0,0.25){\small Çetinaslan};
            \node[darkgray] at  (9.0,0.){\small LMMM };
            \node[darkgray] at  (10.3, 50){\scriptsize .24 };
            \node[darkgray] at  (10.25,42.5){\scriptsize .00 };
            \node[darkgray] at  (10.2, 33.5){\scriptsize .39 };
            \node[darkgray] at  (10.15,25.7){\scriptsize .00 };
            \node[darkgray] at  (10.1, 16.9){\scriptsize .39 };
            \node[darkgray] at  (10.05,9){\scriptsize .00 };
        \end{scope}
        \end{tikzpicture}
  \caption{\textit{Omar}, example frames with predictions using different methods. The odd rows show mesh reconstructions, and regions of higher mesh error are highlighted in red, according to the color bar on the right. The even rows show corresponding blendshape weights activations, with summarized root mean squared error and cardinality of each approach.}
    \label{fig:details_omar}
\end{figure}

The other four datasets lead to similar patterns of the results, further confirming the above conclusions. Numerical results for \textit{Jesse} are given in Figure \ref{fig:Jessebarplots} and Table \ref{tab:tabJesse}, while example frames are in Figures \ref{fig:wide2} and \ref{fig:details_jesse}. For \textit{Vivian}, results are in Figure \ref{fig:Vivianbarplots} and Table \ref{tab:tabVivian}, and example frames are in Figures \ref{fig:wide3} and \ref{fig:details_vivian}. For \textit{Omar}, the results are in Figure \ref{fig:Omarbarplots} and Table \ref{tab:tabOmar}, and example frames are in Figures \ref{fig:wide4} and \ref{fig:details_omar}, while for \textit{Char 5} the results are in Figure \ref{fig:HMbarplots} and Table \ref{tab:tabHM}. We ask readers to look at the supplementary video materials in order to get a better idea of the quality of reconstruction and smoothness of the animated sequences. 

\begin{table}[]
    \centering
    \begin{tabular}{c | c  c  c  c  c  c}
            &\makecell{RMSE \\ mean }&\makecell{RMSE \\ $95^{th}$} & Card.&\makecell{L1 \\ norm}& Rough.         & Time                   \\
                \hline
        Quartic      & 0.025              & 0.077              &\happy\textbf{55.5}& 8.45              &\happy\textbf{0.073}&   5.858             \\
        Linear       & 0.035              & 0.113              &  55.6             &\happy\textbf{8.09}& 0.075              &   0.358             \\        
        Seol         & \done 0.049        & \done 0.147        &  55.6             & 9.74              & 0.082              &   0.064             \\
        Joshi        & 0.019              & 0.074              &  73.8             & \done 11.7        & \done 1.846        &\happy\textbf{0.004} \\
        Çetinaslan   & 0.022              & 0.082              &  72.5             & 10.0              & 0.289              &   0.004             \\
        LMMM         &\happy\textbf{0.012}&\happy\textbf{0.039}& \done 79.5        & 10.1              & 0.280              & \done 14.10
    \end{tabular}
    \caption{\textit{Jesse}. Average values for each metric and each method, corresponding to Figure \ref{fig:Jessebarplots}. The worst score for each column is shaded, while the best is highlighted and bold.}
    \label{tab:tabJesse}
\end{table}

\begin{table}[]
    \centering
    \begin{tabular}{c | c  c  c  c  c  c}
            &\makecell{RMSE \\ mean }&\makecell{RMSE \\ $95^{th}$} & Card.&\makecell{L1 \\ norm}       & Rough.             & Time                   \\
                \hline
        Quartic    & 0.017              & 0.053              &  60.6             &\happy\textbf{10.1}& 0.158              &   5.865          \\
        Linear     & 0.024              & 0.071              &  60.3             & 10.2              & 0.161              &   0.379          \\
        Seol       &  \done 0.050       & \done 0.129        &\happy\textbf{54.9}& 11.2              &\happy\textbf{0.107}&   0.063          \\
        Joshi      & 0.019              & 0.067              &  73.5             & \done 14.4        & \done 2.033        &\happy\textbf{0.004}\\
        Çetinaslan & 0.022              & 0.074              &  73.4             & 11.8              & 0.334              &   0.006          \\
        LMMM       & \happy\textbf{0.013}&\happy\textbf{0.039}& \done 79.5       & 11.9              & 0.324              & \done  14.09
    \end{tabular}
    \caption{\textit{Vivian}. Average values for each metric and each method, corresponding to Figure \ref{fig:Jessebarplots}. The worst score for each column is shaded, while the best is highlighted and bold.}
    \label{tab:tabVivian}
\end{table}

\begin{table}[]
    \centering
    \begin{tabular}{c | c  c  c  c  c  c}
            &\makecell{RMSE \\ mean }&\makecell{RMSE \\ $95^{th}$} & Card.&\makecell{L1 \\ norm}& Rough.         & Time                   \\
                \hline
        Quartic    & 0.023              & 0.065              &\happy\textbf{66.8}& 8.68              & 0.432              & 7.714            \\
        Linear     & 0.035              & 0.116              &  67.9             &\happy\textbf{8.22}& 0.395              & 0.404            \\ 
        Seol       &\done 0.054         &\done 0.160         &  74.2             & 10.1              &\happy\textbf{0.384}& 0.074            \\
        Joshi      & 0.028              & 0.113              &  87.6             & \done 11.1        & \done 4.128        &\happy\textbf{0.004}\\
        Çetinaslan & 0.027              & 0.103              &  88.7             & 8.97              & 0.513              & 0.005            \\
        LMMM       &\happy\textbf{0.013}&\happy\textbf{0.040}& \done 100.        & 9.16              & 0.509              &\done 15.94
    \end{tabular}
    \caption{\textit{Omar}. Average values for each metric and each method, corresponding to Figure \ref{fig:Omarbarplots}. The worst score for each column is shaded, while the best is highlighted and bold.}
    \label{tab:tabOmar}
\end{table}

\begin{table}[]
    \centering
    \begin{tabular}{c | c  c  c  c  c  c}
            &\makecell{RMSE \\ mean }&\makecell{RMSE \\ $95^{th}$} & Card.&\makecell{L1 \\ norm}& Rough.         & Time                              \\
                \hline
        Quartic    & 0.027              & 0.082              &\happy\textbf{53.6}& 7.85              &  0.260             &  1.413              \\
        Linear     & 0.031              & 0.096              &  53.8             &\happy\textbf{7.59}&\happy\textbf{0.228}&  0.072              \\
        Seol       & \done 0.040        & \done 0.110        &  89.1             & 15.1              &  1.536             &  0.011              \\
        Joshi      & 0.015              & 0.065              &  108.             & \done 22.0        & \done    15.40     &  0.004              \\
        Çetinaslan & 0.023              & 0.097              &  107.             & 11.5              &  0.564             &\happy\textbf{0.003} \\
        LMMM       &\happy\textbf{0.017}&\happy\textbf{0.057}& \done 108.        & 11.6              &  0.546             &  \done 4.041
    \end{tabular}
    \caption{\textit{Char 5}. Average values for each metric and each method, corresponding to Figure \ref{fig:HMbarplots}. The worst score for each column is shaded, while the best is highlighted and bold.}
    \label{tab:tabHM}
\end{table}

\subsection{Order of Component Updates}\label{sec:order}

As mentioned earlier in the paper, the order of sequential updates will play an important role in the final results. A strategy that we use is the one proposed in \cite{seol2011artist}, where the blendshapes are ordered according to the magnitude of their offset, as explained in (\ref{eq:ordering}). This strategy is inspired by the artist's intuition since the human animator would start editing by first adjusting more extreme deformers (i.e., blendshapes with higher displacement magnitude) before proceeding to more subtle ones. We refer to this strategy \textit{Decreasing magnitude}, and compare it with several other heuristics. Opposed to that, the blendshapes can be deliberately ordered reverse from (\ref{eq:ordering}) to show that it leads to poor results --- we denote this \textit{Increasing magnitude}. Finally, another simple solution is to order the deformers randomly (\textit{Random ordering}). 

\begin{figure}
    \centering
    \includegraphics[width=0.5\linewidth]{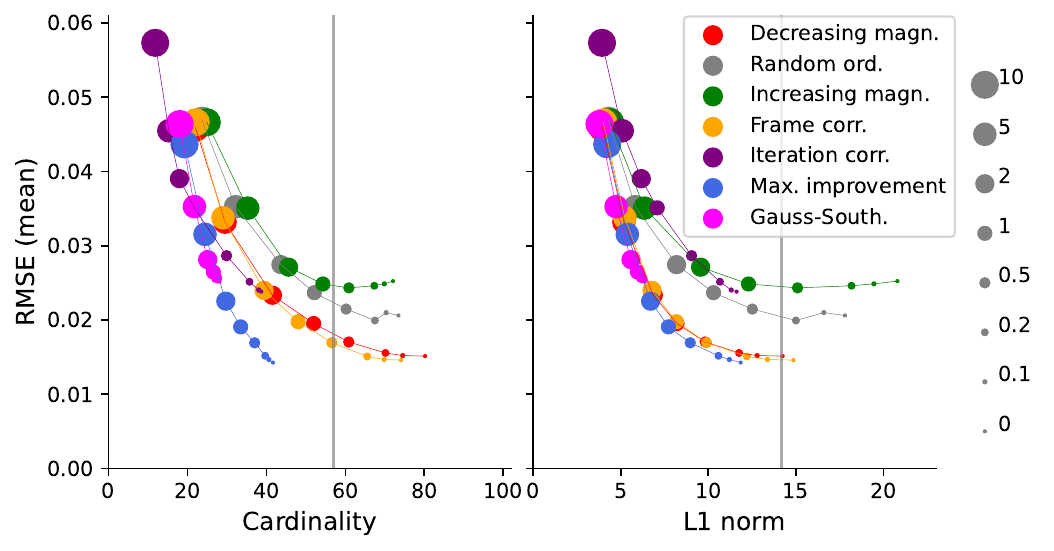}
    \caption{Trade-off between the mesh error and cardinality / L1 norm of the weight vector for different ordering of the updates for \textit{Quadratic} method (\textit{Ada}).}
    \label{fig:order}
\end{figure}

We can also resort to the other tactics that incur a bit higher computational cost (since the order vector needs to be estimated for each frame independently) but might be expected to give a more favorable solution in terms of the objective function. At first, we consider the strategy of Matching Pursuit \cite{mallat1993matching}, where the components are ranked based on the correlation with the target vector. In this sense, we can compute the correlation between each blendshape with the target mesh $\widehat{\textbf{b}}$, and then proceed following that order when fitting the corresponding frame. We call this \textit{Frame correlation}. The other approach is to recompute correlations after each update, that is, when we choose the blendshape with the highest correlation with the target, we estimate its activation weight using (\ref{eq:quartic_solution}), and then recompute the correlations, repeating the process using the updated residual instead of the target mesh. We denote this \textit{Iteration correlation}, and notice that this takes $m$ times more correlation computations per iteration than \textit{Frame correlation}. Another possibility is to use \textit{Gauss-Southwell} update rule \cite{nutini2015coordinate, nutini2017let}, which chooses the coordinate whose derivative at the given point has the largest magnitude. Since we are solving a constrained problem, we need to exclude the candidate weights $w_i$ that have values 0 or 1, such that the gradient indicates unfeasible directions. Finally, we can choose the next component based on the reduction in the objective function, and we will denote it \textit{Maximum improvement}. The idea is to solve (\ref{eq:quartic_solution}) for each weight $w_i$ independently and estimate what would be the value of the cost function (\ref{eq:q_problem}) for each of the updates. Then, we keep only the weight that leads to the biggest decrease of (\ref{eq:q_problem}) and discard all the others. After updating the corresponding weight component, we repeat the process until convergence. The last two approaches might be better than the previous ones in terms of solution fit and sparsity, but they are wasteful in terms of computational costs.

\begin{figure}
    \centering
    \includegraphics[width=0.5\linewidth]{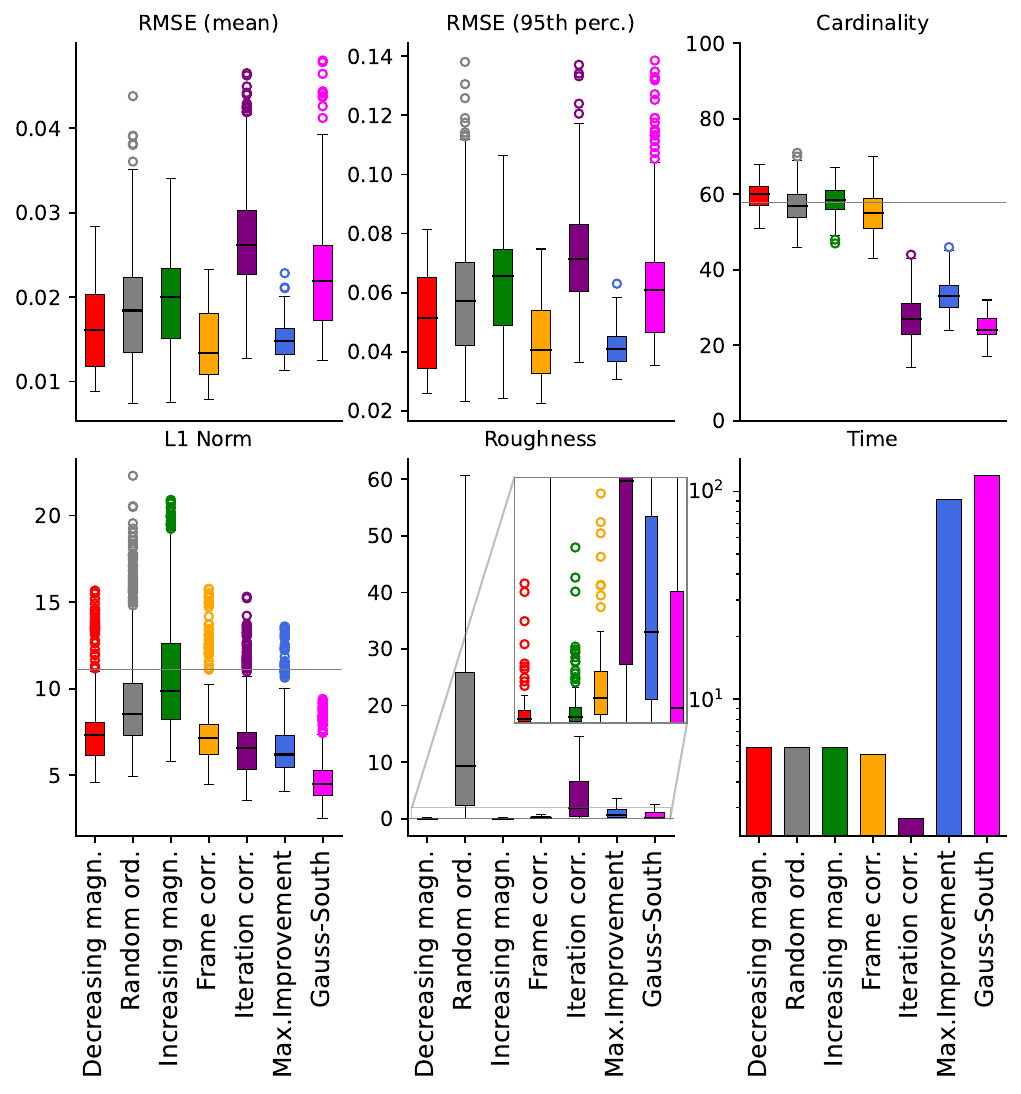}
    \caption{Results statistics for \textit{Ada} for different ordering approaches, over the test animation. Horizontal gray lines indicate the average value of the corresponding metric in the ground truth data. Execution times are presented in the log scale. For the exact numerical values, consult Table \ref{tab:tabOrder}.}
    \label{fig:orderTest}
\end{figure}

The trade-off curves for six variants for \textit{Ada} are presented in Figure \ref{fig:order}. The proposed method of \textit{Decreasing magnitude} has a similar trade-off curve as that of \textit{Frame correlation}, while \textit{Increasing magnitude} and \textit{Random ordering} show both higher error and higher cardinality. \textit{Iteration correlation} leads to slightly lower cardinality at the cost of increased mesh error, while it seems that the \textit{Maximum improvement} approach outperforms the others, while \textit{Gauss-Southwell} partly overlaps with it; but without offering as low RMSE in case of lower regularization. We choose the same value of a regularization parameter $\alpha=0.5$ for all six cases and proceed with the test data. The results are given in Figure \ref{fig:orderTest} and Table \ref{tab:tabOrder}. \textit{Maximum improvement} shows great results in terms of both mesh error and cardinality, however, it is drastically slower than the other methods. The wasteful computations and function evaluations of this approach lead to execution time that is 15 times longer than most other approaches, which makes it unsuitable for practical use. This aspect is similar for \textit{Gauss-Southwell}, while it also yields a relatively high mesh error. The proposed approach of \textit{Decreasing magnitude} is performing relatively well in each aspect, which confirms that it is a good heuristic for this problem. \textit{Frame correlation} behaves similarly to \textit{Decreasing magnitude} --- the main difference is in terms of temporal smoothness, where \textit{Decreasing magnitude} performs better. \textit{Random ordering} is much worse in terms of temporal smoothness than any other approach. This comes as no surprise since, at each frame, the algorithm follows an arbitrary ranking in the fitting phase, hence, even if the individual meshes fit relatively accurately, the consecutive frames will be semantically different. \textit{Increasing magnitude} leads to a bit higher error, but it still gives smooth results, since the controllers are always visited in the same order. Finally, \textit{Iteration correlation} performs poorly in both mesh error and temporal smoothness, even though the produced weight vectors are sparse. 

\begin{table}[]
    \centering
    \begin{tabular}{c | c  c  c  c  c  c}
                             &\makecell{RMSE \\ mean }&\makecell{RMSE \\ $95^{th}$} & Card.            &\makecell{L1 \\ norm} & Rough.              & Time                  \\
                \hline
        \makecell{Decreasing \\ magn.}  & 0.015               & 0.050               & \done 59.3         & 7.68               & \happy\textbf{0.108}& 5.847                \\
    \makecell{Random \\ ord.}           & 0.018               & 0.058               & 57.3               & 9.40               & \done 22.01         & 5.867                \\    
    \makecell{Increasing \\ magn.}      & 0.019               & 0.061               & 58.2               & \done 11.0         & 0.160               & 5.839                \\
    \makecell{Frame \\ corr.}           & \happy\textbf{0.014}& 0.043               & 55.0               & 7.48               & 0.456               & 5.439                \\
    \makecell{Iteration \\ corr.}       & \done 0.026         & \done 0.072         & 27.3               & 6.84               & 4.930               & \happy\textbf{2.654} \\
    \makecell{Maximum \\ Improvement}    & 0.014               & \happy\textbf{0.041}& 32.9               & 6.65               & 1.848               & 91.241               \\
    \makecell{Gauss \\ Southwell}       & 0.022               & 0.061               & \happy\textbf{24.5}& \happy\textbf{4.71}& 0.772               & \done 119.
    \end{tabular}
    \caption{\textit{Ada}. Average values for each metric for different ordering rules, corresponding to Figure \ref{fig:orderTest}. The worst score for each column is shaded, while the best is highlighted and bold.}
    \label{tab:tabOrder}
\end{table}

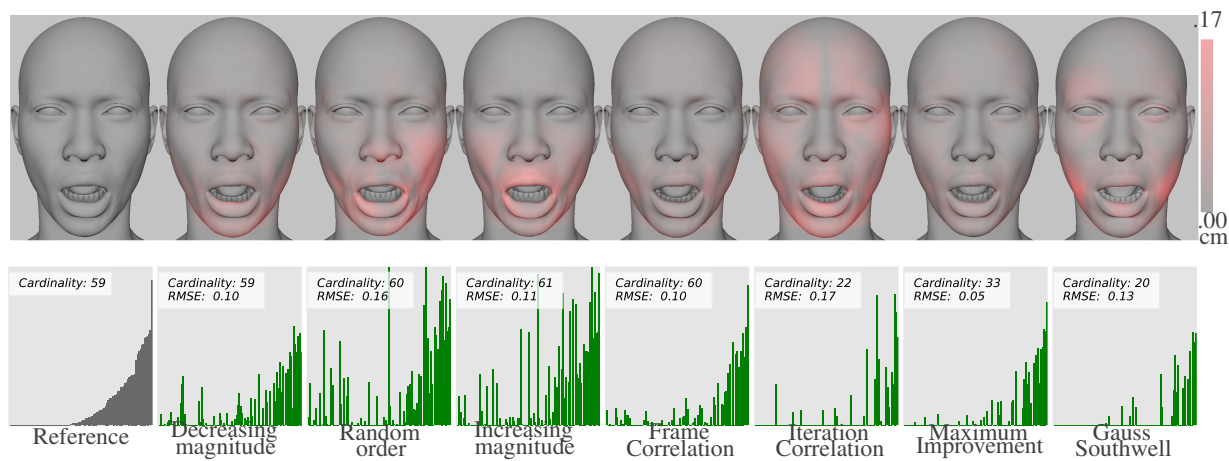
\begin{figure*}
  \centering
  \setlength{\unitlength}{0.1\textwidth}
          \begin{tikzpicture}
        \node[above right, inner sep=0] (image) at (0,2.6){\includegraphics[width=0.968\textwidth]{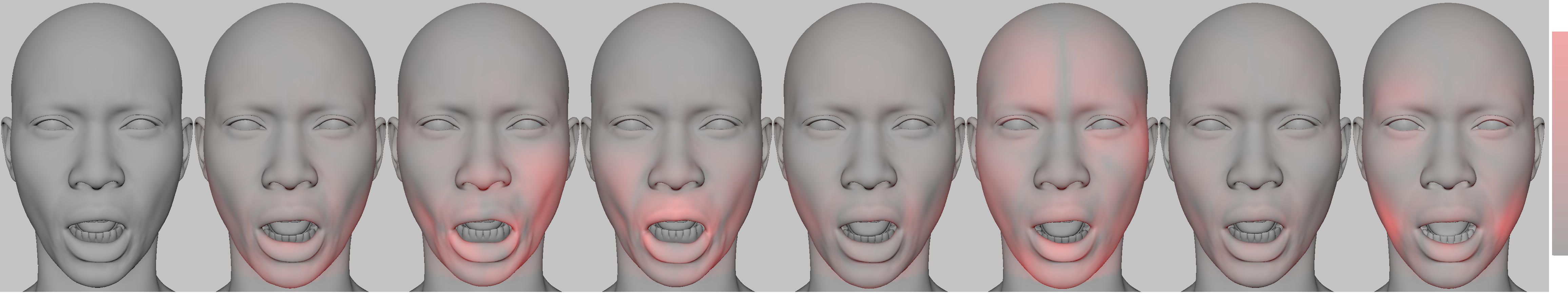}};
        \node[above right, inner sep=0] (image) at (-0.2,0){\includegraphics[width=0.975\textwidth]{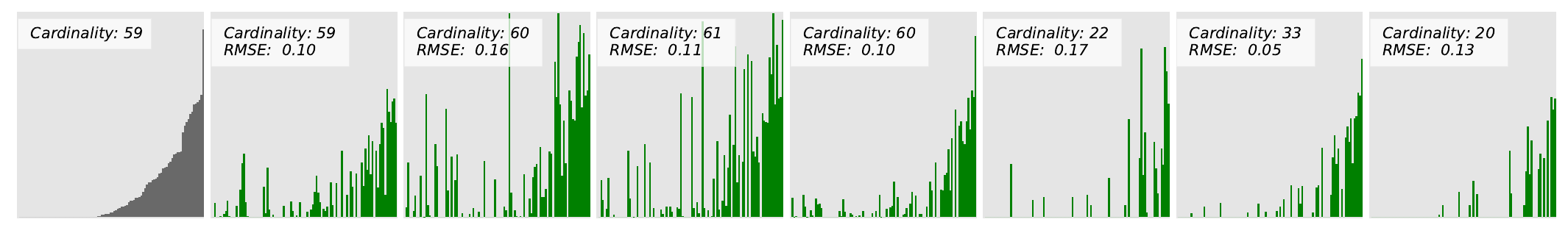}};
        \begin{scope}[
        x={($0.1*(image.south east)$)},
        y={($0.1*(image.north west)$)}]
            \node[darkgray] at  (0.60,0.2){\small Reference };
            \node[darkgray] at  (1.80,0.45){\small Decreasing };
            \node[darkgray] at  (1.80,-0.45){\small magnitude };
            \node[darkgray] at  (3.10,0.45){\small Random };
            \node[darkgray] at  (3.10,-0.45){\small order };
            \node[darkgray] at  (4.30,0.45){\small Increasing };
            \node[darkgray] at  (4.30,-0.45){\small magnitude };
            \node[darkgray] at  (5.60,0.45){\small Frame };
            \node[darkgray] at  (5.60,-0.45){\small Correlation };
            \node[darkgray] at  (6.85,0.45){\small Iteration };
            \node[darkgray] at  (6.85,-0.45){\small Correlation };
            \node[darkgray] at  (8.10,0.45){\small Maximum };
            \node[darkgray] at  (8.10,-0.45){\small Improvement };
            \node[darkgray] at  (9.30,0.45){\small Gauss };
            \node[darkgray] at  (9.30,-0.45){\small Southwell };
            \node[darkgray] at  (10.3,23){\footnotesize .17 };
            \node[darkgray] at  (10.2,12){\footnotesize .00 };
            \node[darkgray] at  (10.2,11){\footnotesize cm };
        \end{scope}
        \end{tikzpicture}
  \caption{\textit{Ada}, an example frame with predictions using different ordering techniques. The top row shows a mesh reconstruction, and regions of higher mesh error are highlighted in red, according to the color bar on the right. The bottom row shows corresponding blendshape weights activation, with summarized root mean squared error and cardinality of each approach.}
  \label{fig:order_meshes}
\end{figure*}

We can also confirm these conclusions by visually inspecting an example frame in Figure \ref{fig:order_meshes}. \textit{Iteration correlation} gives the highest mesh error, with red tones all around the surface, and visible misfits in the mouth region. \textit{Random ordering} and \textit{Increasing magnitude} are slightly better, but the shape of the lips is still different from the reference frame. The other four approaches give good reconstruction, with \textit{Maximum improvement} and \textit{Gauss-Southwell} having far sparser weight vectors than the rest. 

\section{Conclusion}

The method proposed in this paper addresses the inverse rig problem in a coordinate descent manner, using complex nonlinear blendshape models. Our coordinate descent-based method is general enough to work with different levels of corrections or even with a linear rig function. Numerical experiments performed over several datasets show that the proposed method outperforms the state-of-the-art model-based solutions based on the trade-off between mesh error and cardinality of the weight vector, even in the case when using a linear approximation of the rig. Visual inspection further confirms that our method produces a higher-fidelity reconstruction of the original mesh and that it produces a correct facial expression even when the other methods fail. The sequential update rule that is applied implies that our algorithm will not activate mutually exclusive controllers and hence avoid one of the main causes of the instability of a solution. In this respect, the proposed method is somewhat similar to \textit{Seol} \cite{seol2011artist}, yet it is superior in the accuracy of the reconstructed meshes. On the other side, \textit{LMMM} \cite{rackovic2022majorization} gives a high-fidelity mesh reconstruction that sometimes outperforms that of our algorithm, yet it suffers from a high cardinality, similar to \textit{Joshi} \cite{joshi2006learning} and \textit{Çetinaslan} \cite{cetinaslan2020sketching}. For this reason, it is often susceptible to artifacts in the reconstructed meshes and might be hard or impossible to manually adjust later, as opposed to our method. Besides the optimal tradeoff between the mesh error and sparsity of the weights produced by our method, we have seen that our method is three times faster than \textit{LMMM} (the only baseline with comparable mesh reconstruction accuracy) and the resulting animation sequences are visibly more smooth. For all these reasons, our method is a favorable approach in the production of close-shot facial animation, where high accuracy of expression cloning is crucial. 


\textbf{Supplemental Materials}

Supplementary video materials are available at \url{https://youtu.be/Kw4wV24-04c} for \textit{Ada}, \url{https://youtu.be/GLQilLJVI_Q} for \textit{Jesse}, \url{https://youtu.be/1mU6L8MoN4U} for \textit{Omar}, and \url{https://youtu.be/Lo-enH0rfyQ} for \textit{Vivian}. 
For each character we present a reference animation on the left, and a mirrored estimate by different method. Further, the gray bar-plots on the left correspond to weight activations in the reference animation, and the green ones on the right to the estimated solutions. 

\textbf{Funding}

This work has received funding from the European Union's Horizon 2020 research and innovation program under the Marie Skłodowska-Curie grant agreement No. 812912, from FCT IP strategic project NOVA LINCS (FCT UIDB/04516/2020) and project DSAIPA/AI/0087/2018. The work has also been supported in part by the Ministry of Education, Science and Technological Development of the Republic of Serbia (Grant No. 451-03-9/2021-14/200125).

\bibliographystyle{unsrtnat}
\bibliography{references}  






\end{document}